\documentclass[letterpaper]{article} % DO NOT CHANGE THIS
\usepackage{aaai2026}  % DO NOT CHANGE THIS
\usepackage{times}  % DO NOT CHANGE THIS
\usepackage{helvet}  % DO NOT CHANGE THIS
\usepackage{courier}  % DO NOT CHANGE THIS
\usepackage[hyphens]{url}  % DO NOT CHANGE THIS
\usepackage{graphicx} % DO NOT CHANGE THIS
\usepackage{natbib}  % DO NOT CHANGE THIS AND DO NOT ADD ANY OPTIONS TO IT
\usepackage{caption} % DO NOT CHANGE THIS AND DO NOT ADD ANY OPTIONS TO IT
\usepackage{algorithm}
\usepackage{enumitem}
\usepackage{algpseudocode}

\usepackage{longtable}
\usepackage{booktabs}
\usepackage{multirow}
\usepackage{amsthm}
\usepackage{amssymb}
\usepackage{amsmath}
\usepackage{adjustbox}
\usepackage{color}
\usepackage{xcolor}
\usepackage{subfigure}
\usepackage{colortbl} %\cellcolor
\definecolor{Gray}{gray}{0.9}
\definecolor{cvprblue}{rgb}{0.21,0.49,0.74}
\definecolor{deepgreen}{RGB}{0, 190, 0}
\usepackage{bm}
\usepackage{comment}
\newcommand{\vf}{\bm{f}}
\newcommand{\vp}{\bm{p}}

\newcommand{\tophline}{\specialrule{1.5pt}{0pt}{0pt}}
\newcommand{\midhline}{\specialrule{0.75pt}{0pt}{0pt}}
\newcommand{\bottomline}{\specialrule{1.5pt}{0pt}{0pt}}
\usepackage{newfloat}
\usepackage{listings}
\DeclareCaptionStyle{ruled}{labelfont=normalfont,labelsep=colon,strut=off} % DO NOT CHANGE THIS
\floatstyle{ruled}
\newfloat{listing}{tb}{lst}{}
\floatname{listing}{Listing}
\title{Heterogeneous Complementary Distillation}

\usepackage{rotating}
\author{
    Liuchi Xu\textsuperscript{\rm 1,\rm 4}, Hao Zheng\textsuperscript{\rm 2}, Lu Wang\footnotemark[1]\textsuperscript{\rm 1}, Lisheng Xu\textsuperscript{\rm 3}, Jun Cheng\thanks{Corresponding authors}\textsuperscript{\rm 4,\rm 5}\\
}
\affiliations{
    \textsuperscript{\rm 1}School of
Computer Science and Engineering, Northeastern University, China; \\
\textsuperscript{\rm 2}School of Computer Science, South China Normal University, China;\\
 \textsuperscript{\rm 3}College of Information Science and Engineering, Northeastern University, China;\\
\textsuperscript{\rm 4}Guangdong-Hong Kong-Macao Joint Laboratory of Human-Machine Intelligence-Synergy Systems,\\Shenzhen Institutes of Advanced Technology, Chinese Academy of Sciences, China;\\
\textsuperscript{\rm 5}The Chinese University of Hong Kong, Hong Kong, China\\
    \{xuliuchi@stumail,wanglu@mail\}.neu.edu.cn, \
zzeo.zheng@gmail.com, \\
xuls@mail.neu.edu.cn, \
Jun.cheng@siat.ac.cn
}

\usepackage{bibentry}
\begin{document}

\maketitle

\begin{abstract}
Knowledge distillation (KD) transfers the ``dark knowledge'' from a complex teacher model to a compact student model. However, heterogeneous architecture distillation, such as Vision Transformer (ViT) to ResNet18, faces challenges due to differences in spatial feature representations. Traditional KD methods are mostly designed for homogeneous architectures and hence struggle to effectively address the disparity. Although heterogeneous KD approaches have been developed recently to solve these issues, they often incur high computational costs and complex designs, or overly rely on logit alignment, which limits their ability to leverage the complementary features. To overcome these limitations, we propose Heterogeneous Complementary Distillation (HCD), a simple yet effective framework that integrates complementary teacher and student features to align representations in shared logits. These logits are decomposed and constrained to facilitate diverse knowledge transfer to the student. Specifically, HCD processes the student’s intermediate features through convolutional projector and adaptive pooling, concatenates them with teacher's feature from the penultimate layer and then maps them via the Complementary Feature Mapper (CFM) module, comprising fully connected layer, to produce shared logits.  We further introduce Sub-logit Decoupled Distillation (SDD) that partitions the shared logits into $n$ sub-logits, which are fused with teacher's logits to rectify classification. To ensure sub-logit diversity and reduce redundant knowledge transfer, we propose an Orthogonality Loss (OL). By preserving student-specific strengths and leveraging teacher knowledge, HCD enhances robustness and generalization in students. Extensive experiments on the CIFAR-100, Fine-grained (e.g., CUB200, Aircraft) and ImageNet-1K datasets demonstrate that HCD outperforms state-of-the-art KD methods, establishing it as an effective solution for heterogeneous KD. Code is available at: https://github.com/yema-web/HCD

\end{abstract}

%\begin{links}
%    \link{Code}%{https://github.com/yema-web/HCD}
%\end{links}

\section{Introduction}

Knowledge distillation (KD), proposed by Hinton et al.~\cite{hinton2015distilling}, aligns the soft label distributions of a complex teacher model (abbreviated teacher) and a compact student model (abbreviated student) using Kullback-Leibler (KL) divergence~\cite{kullback1951information}. This technique enables the student's performance to approach that of the teacher's. Due to its simplicity and its ability to preserve the original architecture, KD is widely applied to resource-constrained computer vision tasks, such as image classification~\cite{he2016deep,yang2022cross,zheng2025hierarchical}, object detection~\cite{li2024detkds,zhao2024detrs,li2023kd} and semantic segmentation~\cite{zou2024segment,liang2023open}. However, traditional variants of KD, such as FitNets~\cite{romero2014fitnets}, DKD~\cite{zhao2022decoupled}, and RKD~\cite{park2019relational}, mainly focus on distillation between homogeneous architectures (e.g., ResNet34 to ResNet18). This limits the improvement of the student's performance due to the lack of knowledge diversity in the KD strategy. In contrast, Vision Transformers (ViTs)~\cite{dosovitskiy2020image,tolstikhin2021mlp}, with stronger representational capacity, provide new avenues for heterogeneous architecture (or cross-architecture) distillation~\cite{liu2022cross} and allow the student to acquire diverse feature representations from the teacher. Intuitively, heterogeneous architecture distillation, such as ViT-to-ResNet18, leverages the complementary information of inductive biases from the teacher and the student. As a result, it enhances the student's robustness and generalization beyond the limitations of homogeneous distillation methods. However, when traditional feature-based distillation methods~\cite{romero2014fitnets,park2019relational} in homogeneous settings are applied to heterogeneous settings, distillation result often fails due to mismatched spatial feature representations (e.g., differences in receptive field size), which results in the degradation of the student's performance.\\
\indent To overcome the heterogeneous feature representation gap, existing methods, such as OFA-KD~\cite{hao2023one}, adopt a one-for-all strategy at each student stage to map intermediate representations to the logits space and align them with the teacher’s logits. However, relying solely on logits alignment between teacher and student fails to leverage their complementary strengths, thus limiting potential performance gains. Similarly, PAT proposes a region-aware attention (RAA) module~\cite{lin2025feature} that restructures student intermediate features to align with teacher representations, improving student's performance but incurring high computational costs and complex designs. These limitations highlight the challenge of transferring heterogeneous knowledge effectively, primarily due to the complex spatial feature patterns that students struggle to fully capture in heterogeneous distillation. In particular, over-reliance on the teacher’s logits, as seen in OFA-KD, may cause the student to overlook its own feature extraction strengths. Therefore, our research focuses on tailoring logit knowledge to integrate the feature representations from both teacher’s and the student’s architectural strengths, facilitating more effective knowledge transfer in heterogeneous distillation.
\begin{comment}
\begin{figure}[!t]
	\centering
	 \subfigure[CNN/CNN] {\includegraphics[width=.15\textwidth]{AnonymousSubmission/LaTeX/cka_similarity_cnn_cnn.png}}
	\subfigure[ViT/ViT] {\includegraphics[width=.15\textwidth]{AnonymousSubmission/LaTeX/cka_similarity_vit_vit.png}}
    \subfigure[CNN/ViT] {\includegraphics[width=.15\textwidth]{AnonymousSubmission/LaTeX/cka_similarity_cnn_vit.png}}
 \vspace{-1.2em}
	\caption{Similarity heatmap of intermediate features measured by CKA~\cite{cortes2012algorithms,kornblith2019similarity}, using ResNet101 and ViT-Small.}
	\label{heatmap}
 \vspace{-20pt}
\end{figure}
\end{comment}

To this end, we propose \textbf{H}eterogeneous \textbf{C}omplementary \textbf{D}istillation (HCD), a novel approach that leverages complementary teacher-student feature representations to map them into a shared logits space, where logits are decomposed and constrained for diverse knowledge transfer to the student. Specifically, outputs from the student’s intermediate $i$-th stages are processed through a convolutional module and adaptive pooling, and then concatenated with the teacher’s penultimate layer features. The concatenated features are then fed into the \textbf{C}omplementary \textbf{F}eature \textbf{M}apper (CFM) module, comprising fully connected layer, to produce shared logits output that integrates teacher and student spatial representations while preserving student-specific features. However, the direct learning of shared logits knowledge presents challenges, as they contain global contextual information that is difficult for the student to grasp. To solve this problem, we propose \textbf{S}ub-logit \textbf{D}ecoupled \textbf{D}istillation (SDD). It decomposes shared logits into $n$ sub-logits, which are then fused with the teacher’s logits to ensure classification consistency. Furthermore, to promote diversity and non-redundant sub-logit representations, we introduce an  \textbf{O}rthogonality  \textbf{L}oss (OL) to ensure the $n$ sub-logits capture distinct characteristics. The HCD approach enables the student to effectively assimilate the teacher’s feature knowledge while preserving its own strengths, thereby improving robustness and generalization. Extensive experiments on datasets, such as CIFAR-100, Fine-grained (e.g., CUB-200-2011, FGVC-Aircraft), and ImageNet-1K, have confirmed the effectiveness of our HCD method. We summarize the contributions of this paper as follows:

\begin{itemize}
    \item We design a Complementary Feature Mapper (CFM) to produce shared logits space, which addresses the disparate features between the teacher and the student.
    \item We propose Sub-logit Decoupled Distillation (SDD) to decompose shared logits space and obtain various sub-logit, facilitating more effective and specialized knowledge transfer to the student.
    \item We introduce an orthogonal loss to emphasize sub-logit diversity, thereby improving the efficiency of knowledge transfer by promoting non-redundant feature learning.
    \item Extensive experiments on CIFAR-100,  Fine-grained, and ImageNet-1K datasets consistently demonstrate that our proposed method outperforms current state-of-the-art KD
    methods, achieving superior performance.
\end{itemize}

\section{Related Work} \label{Related Work}
 In this section, we review existing homogeneous and heterogeneous knowledge distillation approaches, highlighting their differences below.
 
\textbf{Homogeneous Architecture Knowledge Distillation.} Knowledge distillation (KD) transfers the teacher's ``dark knowledge'' to a compact student, enabling the student to achieve performance comparable to the teacher's while reducing the model size. KD methods are broadly categorized into feature-based KD~\cite{li2023curriculum,yang2022focal,zhao2024detrs,romero2014fitnets,liu2023norm,tian2019contrastive,li2023automated,chen2021distilling,liu2023functionconsistent,dong2023diswot,guo2023class,huang2024knowledge,kim2024do,yang2022masked,yang2021knowledge,wang2019distilling,li2022shadow} and logit-based  KD~\cite{zhao2022decoupled,yang2023knowledge,li2022asymmetric,huang2022knowledge,xu2024improving,luo2024scale,Auxiliary,zheng2024knowledge,Sun2024Logit,jin2023multi}. Feature-based KD methods use the teacher’s intermediate features to improve the student’s feature representations. For example, FitNets~\cite{romero2014fitnets} aligns intermediate features across stages using mean squared error.
NORM~\cite{liu2023norm} enables efficient many-to-one feature matching via expanded student representations.
CRD~\cite{tian2019contrastive} incorporates contrastive learning to improve knowledge transfer efficiency. ReviewKD~\cite{chen2021distilling} introduces a review learning mechanism that improves the efficiency of knowledge transfer from the teacher. Despite their effectiveness, these methods often incur high computational and memory cost, prompting a shift toward logit-based distillation approaches. For instance, DKD~\cite{zhao2022decoupled} decouples logit knowledge into target and non-target class KD. LDRLD~\cite{xu2025local} mines fine-grained logit relationships to improve the student's performance. IKD~\cite{wang2025debiased} uses intra-class distillation to mitigate prediction biases within the same class. WTTM~\cite{zheng2024knowledge} refines the temperature scaling in teacher matching. RLD~\cite{sun2024knowledge} uses
labeling information to dynamically refine teacher knowledge. TeKAP~\cite{hossain2025single} employs multi-perspective teacher knowledge via random sampling. Following the success of KD methods in homogeneous architecture, researchers have shifted to address architectural disparities between the teacher and the student.

\textbf{Heterogeneous Architecture Knowledge Distillation.} Architectural disparities between the teacher and student often limit the effectiveness of homogeneous distillation methods in heterogeneous settings. To tackle this challenge, researchers have proposed various methods~\cite{hao2023one,li2024tas,lin2025feature,zhao2023cumulative,zhou2025all,huang2025distilling,lee2025customkd,zhang2025cross}. For example, MLDR-KD~\cite{yang2025multilevel} proposes a novel heterogeneous relational KD framework that retains dark knowledge while boosting confidence in the correct target.  
RSD~\cite{zhang2025cross} extracts architecture-agnostic knowledge by reducing redundant architecture-specific information. PAT~\cite{lin2025feature} introduces a region-aware attention (RAA) module that aligns teacher representations by restructuring multi-layer student features. LFCC~\cite{wu2024aligning} uses a multi-scale low-pass filter and its learnable derivative to compress low-frequency components of teacher and student features, reducing representational disparity and mitigating the impact of spatial noise. TCS~\cite{zhou2025all} tailors a coordinate system to transfer ``dark knowledge" from a task-agnostic teacher to task-specific student networks.  HeteroAKD~\cite{huang2025distilling} transforms heterogeneous knowledge into the logit space, mitigating architecture-specific influences. CustomKD~\cite{lee2025customkd} customizes the generalized features of large vision foundation models (LVFMs) for the student, minimizing spatial representation differences. However, methods like HeteroAKD, LFCC, and PAT, which rely on all intermediate stages, incur high computational costs and decrease the efficiency of knowledge transfer. Consequently, heterogeneous KD approaches require further optimization to overcome these limitations. To address these challenges, our proposed method integrates complementary feature mapping from the teacher and student, sub-logit decoupled distillation, and orthogonal constraints to enhance efficiency and effectiveness.
\begin{figure*}[!ht]
\center{\includegraphics[width=17.50cm]{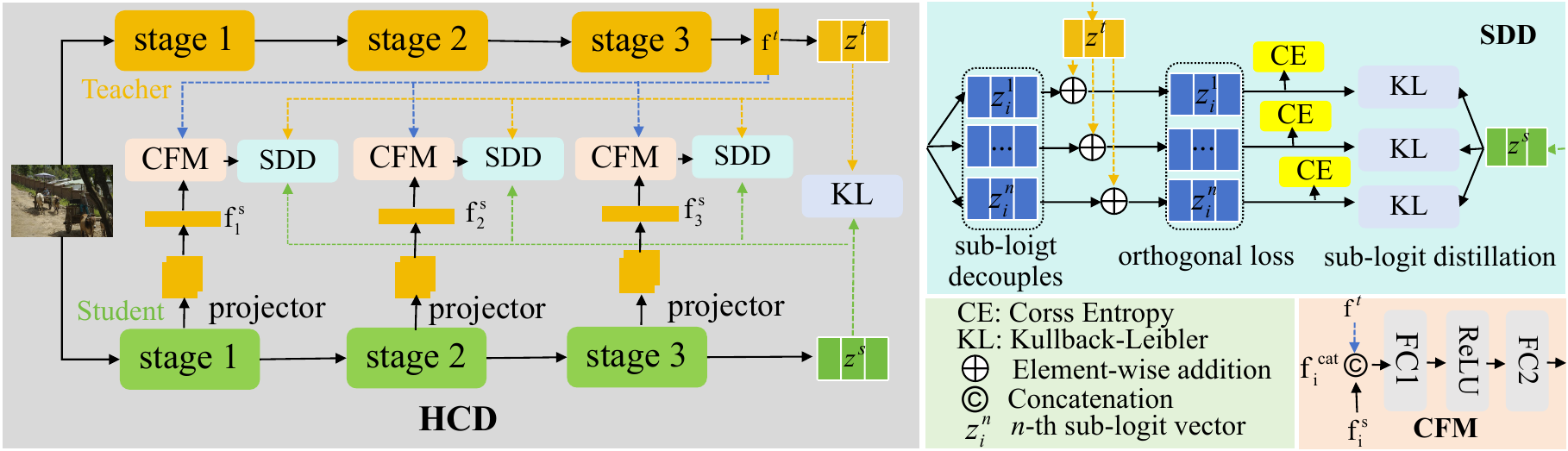}} 
 \caption{
  \label{framwork}
      Overview of the proposed HCD framework, which includes the three components.} 
\end{figure*}
\section{Method}
In this section, we first review knowledge distillation (KD) methods and our motivation. We then propose Heterogeneous Complementary Distillation (HCD) for heterogeneous KD. Finally, we optimize the total loss function.
\subsection{Preliminary}
\label{knowledge distillation}
 KD was originally proposed by Hinton et al.~\cite{hinton2015distilling}. By minimizing the output probability distributions of the teacher and the student via KL divergence, KD transfers the complex teacher's ``dark knowledge" to the compact student. Specifically, the KD loss, $\mathcal{L}_{KL}$, uses KL divergence to measure the difference between the soft label distributions of the student and the teacher, while the cross-entropy loss, $\mathcal{L}_{CE}$, supervises the student’s classification performance. The total distillation loss combines two terms as follows:
\begin{equation}
\begin{aligned}
\mathcal{L}_{total} &= \alpha \mathcal{L}_{CE}(\sigma(\mathbf{z}^{s}),\mathbf{y})  + (1-\alpha ) \mathcal{L}_{KL}(\vp^t||\vp^s)\\
&=-  \alpha \times \mathbf{y} \log \sigma(\mathbf{z}^{s}) + (1-\alpha ) \vp^t \log ( \vp^t/\vp^s),
\end{aligned}
\end{equation}
where $\vp^{s} = \sigma(\mathbf{z}^{s}/\tau)$ and $\vp^t =  \sigma(\mathbf{z}^t/\tau)$ denote the student's and the teacher's soft label distributions derived from their logits $\mathbf{z}^s$ and $\mathbf{z}^t$ scaled by temperature $\tau$, $\sigma \left( \cdot\right)$ denotes the softmax function. The vector $\mathbf{y}$ represents the one-hot labels, and the parameter $\alpha$ balances the two loss terms.

\subsection{Motivation}
\label{motivation}
Heterogeneous architectures, such as convolutional neural networks (CNNs), vision transformers (ViTs), and MLP-Mixers, generate distinct spatial feature representations due to their inherent inductive biases. For instance, CNNs emphasize locality and translation invariance, whereas ViTs excel at modeling global contextual relationships. Direct alignment of the intermediate feature representations using FitNets~\cite{romero2014fitnets} between a student (e.g., CNN) and a teacher (e.g., ViT) often results in failure due to distinct representations. This consequently hinders effective knowledge transfer and degrades student performance. Intuitively, heterogeneous distillation can enrich knowledge diversity by leveraging differences in feature representation between architectures, thereby improving the student's performance. However, traditional homogeneous distillation methods, such as feature alignment in FitNets or logit distillation in KD, fail to address these representation disparities, leading to suboptimal student performance. As a result, they cannot fully exploit the inductive bias information of each architecture to enhance the student's performance. While existing heterogeneous distillation methods attempt to address this issue, they still have limitations. For example, OFA’s single-stage logit mapping oversimplifies the process, limiting the student’s feature representation capacity. Conversely, PAT’s RAA mitigates feature representation misalignment effectively but incurs significant computational overhead. To address these issues, The proposed \textbf{H}eterogeneous \textbf{C}omplementary \textbf{D}istillation (HCD) method. The HCD integrates the teacher’s deep feature representations with the student’s intermediate feature representations and enhances the student's performance.
\subsection{Heterogeneous Complementary Distillation (HCD)}
An overview of the proposed HCD framework is illustrated in Fig.~\ref{framwork}, including CFM, SDD, and orthogonal loss.
\subsection{Complementary Feature Mapper}
\label{cfm module}
To address the challenge of aligning intermediate layer feature in the heterogeneous architecture KD, caused by differences in inductive biases, we propose a simple yet effective module designated as the \textbf{C}omplementary \textbf{F}eature \textbf{M}apper (CFM). The CFM concatenates the features from different intermediate stages of the student (e.g., 
$i$-th stage output) with the penultimate layer features of the teacher and maps them into shared logits space. Specifically, given a batch $x$ with $B$ samples for a ViT teacher, we extract the feature from the penultimate layer of the teacher's encoder $\mathcal{F}^t$, denoted as $\vf^t =\mathcal{F}^t(x) \in \mathbb{R}^{B \times d}$, where $d$ represents the feature dimension (e.g., $d=1024$ for ViT-L). For a CNN student, we extract the intermediate feature from the student's encoder $\mathcal{F}^{s}$, denoted as $\mathcal{G}_{i}^{s} =\mathcal{F}^{s}_i(x) \in \mathbb{R}^{B \times C_i \times H_i \times W_i}$ at the $i$-th stage output of the student, where $C_i$, $H_i$, and $W_i$ represent the number of channel, height, and width, respectively. The intermediate feature $\mathcal{G}_{i}^{s}$ is fed into a pair of convolutional blocks, each consisting of a $3 \times 3$ convolution (Conv), batch normalization (BN), and ReLU function, followed by adaptive average pooling (Pool) to yield $\vf_i^{s} \in \mathbb{R}^{B \times m}$, where $m$ is the feature dimension (e.g., $m=256$), is formulated as follows:
\begin{equation}
\begin{aligned}
\vf_i^{s}=\operatorname{Pool}\left(2\times\operatorname{ReLU}\left(\operatorname{BN}\left(\operatorname{Conv}\left(\mathcal{G}_{i}^{s}\right)\right)\right)\right)\in \mathbb{R}^{B\times m}.
\end{aligned}
\end{equation}
To effectively exploit the complementary inductive biases of heterogeneous architectures, we propose the CFM module. In particular, we concatenate the mapped features $\vf_i^{s}$ from the $i$-th stage of the student with the penultimate layer features of the teacher $\vf^{\text {t}}$, as formulated below:
\begin{equation}
\begin{aligned}
\quad \vf^{\text {cat }}_i=\operatorname{Concat}\left(\vf_i^{s},\vf^t\right)\in \mathbb{R}^{B\times(d+m)},
\end{aligned}
\end{equation}
where $\operatorname{Concat}(\cdot)$ denotes the concatenation operation along the feature dimension. We then project the concatenated features, $\vf^{\text{cat}}_{i}$, through fully connected layer (FC), such as two layers,  to map them into the shared logits space, $\mathbf{z}_i$, as expressed below:
\begin{equation}
\begin{aligned}
\mathbf{z}_i=\operatorname{FC}_2\left(\operatorname{ReLU}\left(\operatorname{FC}_1\left(\vf^{\text {cat }}_i\right)\right)\right)\in \mathbb{R}^{B\times K},
\end{aligned}
\label{cfm}
\end{equation}
where $\operatorname{FC}_1: \mathbb{R}^{d+m} \to \mathbb{R}^d$ and $\operatorname{FC}_2: \mathbb{R}^{d} \to \mathbb{R}^K$, and $K$ represents the number of classes. As a result, this process can reduce the information gap between the teacher and the student. The proposed module effectively mitigates misalignment between spatial feature representations and leverages the complementary inductive biases of the teacher and the student to improve performance. 

In this way, CFM effectively leverages the strengths of both the teacher and the student in two key aspects. First, CFM integrates low-level feature maps from the student (e.g., edges and textures in the CNN) with high-level semantic features from the teacher (e.g., class-discriminative features in the ViT), thus enhancing the diversity of feature representations. Second, CFM combines the complementary inductive biases of heterogeneous architectures, such as the locality and translation invariance in CNN student and the global contextual relationships with attention mechanism of the ViT teacher, to generate more comprehensive feature representations. Consequently, CFM improves the effectiveness of heterogeneous knowledge distillation and enhances the robustness of the student.
\subsection{Sub-logit Decoupled Distillation}
\label{sub-logit}
\textbf{Sub-logit Decomposition:} Directly aligning the student’s logits with the CFM-mapped shared logits presents a challenge for optimization, as these logits incorporate global contextual representations that are difficult for the student to learn. To address this, we extend the CFM’s $\operatorname{FC}_2$ output dimension from $\mathbb{R}^{B \times K}$ to $\mathbb{R}^{B\times (n \times K)}$ in Eq.~(\ref{cfm}), and refer to it as $\overline{\operatorname{FC}_2}$. Then, we decompos the shared logits output into $n$ sub-logits tailored to the student’s representation capacity. Such decomposition enables the student to effectively capture the complementary knowledge. The decomposition process is defined as follows:
\begin{equation}
\begin{aligned}
\mathbf{z}_i=\overline{\operatorname{FC}_2}\left(\operatorname{ReLU}\left(\operatorname{FC}_1\left(\vf^{\text {cat }}_i\right)\right)\right)\in \mathbb{R}^{B\times (n\times K)},
\end{aligned}
\label{cfm1}
\end{equation}
where $\mathbf{z}_i = \left[\mathbf{z}_{i}^{1}, \mathbf{z}_{i}^{2}, \dots, \mathbf{z}_{i}^{n} \right]$, and $\mathbf{z}_{i}^{j}\in \mathbb{R}^{B\times K}$, $1\leq j\leq n$.\\  
\textbf{Sub-logit Augmentation:} To ensure each sub-logit $\mathbf{z}_{i}^{j}$ remains consistent with the teacher's predictions and to prevent deviation from the task objective, we fuse each sub-logit with the teacher's logits $\mathbf{z}^t$,  expressed as follows:
\begin{equation}
\begin{aligned}
\mathbf{z}_{i}^{j} \gets \mathbf{z}_{i}^{j} \oplus  \mathbf{z}^t, \quad  \text{for each} \quad j=1,2, \ldots, n,
\end{aligned}
\label{fusion}
\end{equation}
where the symbol $\oplus$ denotes element-wise addition. The fusion update of each sub-logit $\mathbf{z}_{i}^{j}$ is concatenated column-wise to obtain the final decomposed logits \( \mathbf{z}_i^{\prime} \) as follows:
\begin{equation}
\begin{aligned}
\mathbf{z}_i^{\prime} = \left[ \mathbf{z}_{i}^{1}, \mathbf{z}_{i}^{2}, \dots, \mathbf{z}_{i}^{n} \right] \in \mathbb{R}^{B \times (n \times K)}.
\end{aligned}
\label{delogit}
\end{equation}
\textbf{Sub-logit Knowledge Transfer:} The goal of sub-logit is to enable the student to fully acquire the complementary knowledge from the teacher. Thus, we use KL divergence ($\mathcal{L}_{\text{KL}}$) to transfer decomposed logits knowledge from $\mathbf{z}^{\prime}_{i}$ to the student's $\mathbf{z}^s$, and ensure accurate sub-logit classification using cross-entropy loss ($\mathcal{L}_{\text{CE}}$). Note that instead of aligning the teacher’s raw logits $\mathbf{z}^t$, we aim to preserve the student’s knowledge while incorporating the teacher’s heterogeneous information to enhance comprehension. The sub-logit losses, $\mathcal{L}_{KL}^{sub}$ and $\mathcal{L}_{CE}^{sub}$, are expressed as follows:
\begin{equation}
\begin{aligned}
\mathcal{L}_{KD}^{sub} = \frac{1}{l\times n}  \sum_{i=1}^{l}\sum_{j=1}^{n}\mathcal{L}_{KL}(\vp_{i}^j||\vp^s),  %\quad  \text{for each} \quad j=1,2, \ldots, n, \text{and}\quad l=1,2, \ldots, i,
\end{aligned}
\label{kd}
\end{equation}
\begin{equation}
\begin{aligned}
\mathcal{L}_{CE}^{sub} = \frac{1}{l\times n}\sum_{i=1}^{l}\sum_{j=1}^{n}\mathcal{L}_\text{CE}(\sigma(\mathbf{z}^{j}_{i}),\mathbf{y}), %\quad  \text{for each} \quad j=1,2, \ldots, n, \text{and}\quad l=1,2, \ldots, i,
\end{aligned}
\label{ce_loss}
\end{equation}
where $\vp^s$ = $\sigma(\mathbf{z}^s/\tau)$ and $\vp^{j}_{i}$ = $\sigma(\mathbf{z}^{j}_{i}/\tau)$ denote the student's and sub-logit's output probability distributions, respectively, which are obtained by normalizing their logits $\mathbf{z}^s$ and $\mathbf{z}_i^{\prime}$, with the batch dimension $B$ omitted for simplicity. $l$ is the total number of stages of the student.
\subsection{Orthogonal Loss for Sub-logit Diversity}
\label{diversity}
\textbf{Masking Ground-truth Label Position:} According to Eq.~(\ref{ce_loss}), cross-entropy loss enforces uniform probability distributions of sub-logits relative to the ground-truth label, leading to similar predictions for the same label across all sub-logits. To alleviate this problem
and promote diversity among the sub-logits $\mathbf{z}_i^j$, while excluding the influence of ground-truth label position prediction, we modify the sub-logit by suppressing the value at the ground-truth label index $Y \in \{0,1,\ldots,K-1\}$ for each sub-logit. 
\begin{comment}
The modified sub-logit vector, $\mathbf{z}_i^j \in \mathbb{R}^K$, is defined as follows:
\vspace{-4pt}
\begin{equation} 
\begin{aligned}
\mathbf{z}_{i,k}^j =
\begin{cases} 
-\epsilon & \text{if } \text{Index}(k) = Y, \\
\mathbf{z}_{i,k}^j & \text{otherwise},
\end{cases}
\end{aligned}
\vspace{-6pt}
\end{equation} 
where $\mathbf{z}_{i,k}^j$ is the $k$-th element of $\mathbf{z}_i^j$, Index($k$) denotes the index of element $k$, and $\epsilon = 10^{-6}$. 
\end{comment}
We adjust the sub-logit vector, $\mathbf{z}_i^j \in \mathbb{R}^K$, using a mask-based approach as follows:
\begin{equation}
\begin{aligned}
\mathbf{z}_i^j \leftarrow \mathbf{z}_i^j \odot (1 - \mathbf{m}_i) - \epsilon \cdot \mathbf{m}_i,
\end{aligned}
\label{masking}
\end{equation}
where the symbol $\odot$ denotes element-wise multiplication, $\mathbf{m}_i \in \{0,1\}^K$ is a one-hot vector, where $\mathbf{m}_{i,k} = \delta_{k,Y}$, and $\delta_{k,Y} = 1$ if $k = Y$, otherwise 0, $k$ represents the index of the maximum value of $\mathbf{z}_i^j$, and $\epsilon = 10^{-6}$ prevents zero values in the ground-truth labels to ensure numerical stability. This modification ensures that the ground-truth label index does not contribute to the diversity loss, enabling effective diversification of non-target class predictions across sub-logits.\\
\textbf{Orthogonal Loss:} To fully utilize the discriminative information among sub-logits while reducing their similarity, we impose orthogonal constraints to promote the diversity of the sub-logits, thereby enabling the student to effectively acquire the teacher’s global spatial feature representation. Following Eq.~(\ref{delogit}) and Eq.~(\ref{masking}) that $\mathbf{z}_i^{\prime} = \left[ \mathbf{z}_{i}^{1}, \mathbf{z}_{i}^{2}, \dots, \mathbf{z}_{i}^{n} \right]\in \mathbb{R}^{n\times K}$, with the batch dimension $B$ omitted for simplicity, each sub-logit $\mathbf{z}_i^j$ is normalized to obtain its normalized vector $\bar{\mathbf{z}}_i^j$, which is defined as:
\begin{equation} 
\begin{aligned}
\bar{\mathbf{z}}_i^j=\frac{\mathbf{z}^{j}_{i}}{||\mathbf{z}^{j}_{i}||_2},\quad \left\| \mathbf{z}_i^j \right\|_2 = \sqrt{\sum_{k=1}^{K} \left( \mathbf{z}_{i,k}^j \right)^2}, %\quad  \text{for each} \quad j=1,2, \ldots, n,
\end{aligned}
\end{equation} 
where the vector $\bar{\mathbf{z}}_i^j$ satisfies $||\bar{\mathbf{z}}_i^j||_2$ = 1. Next, we compute the dot product of any two sub-logits $\bar{\mathbf{z}}_i^p$ and $\bar{\mathbf{z}}_i^q$, denoted by the matrix $\mathcal{A}_{pq}$ as follows:
\begin{equation} 
\begin{aligned}
\mathcal{A}_{p q}=\bar{\mathbf{z}}_i^p \cdot \bar{\mathbf{z}}_i^q = \frac{\mathbf{z}^{p}_{i}}{||\mathbf{z}^{p}_{i}||_2} \cdot \frac{\mathbf{z}^{q}_{i}}{||\mathbf{z}^{q}_{i}||_2},
\end{aligned}
\end{equation} 
where $1\leq q \leq n$ and $1\leq p \leq n$. For $p$ = $q$, the diagonal elements represent the self-correlation, where $\mathcal{A}_{p q}$ = 1. To calculate the dot product between different vectors, we exclude the diagonal elements of the matrix $\mathcal{A}_{p q}$, obtaining a new matrix $\mathcal{A}^{\prime}$, which contains all the off-diagonal elements:
\begin{equation} 
\begin{aligned}
\mathcal{A}^{\prime}=\left\{\mathcal{A}_{p q} \mid p \neq q\right\}.
\end{aligned}
\end{equation} 
The orthogonal loss function computes the mean squared value of the off-diagonal elements of the matrix $\mathcal{A}^{\prime}$. We enforce these off-diagonal elements to be small, ensuring that distinct sub-logits remain orthogonal. The orthogonal loss, $\mathcal{L}_{orth}$, is formulated as follows:
\begin{equation} 
\adjustbox{scale=0.99}{$
\begin{aligned}
\mathcal{L}_{orth
}=\frac{1}{i\times n(n-1)} \sum_{i}\sum_{p \neq q} \left[ \max(0, \mathcal{A}^{\prime} - \theta) \right]^2,
\end{aligned}
$}
\label{orth}
\end{equation} 
where $\theta$ serves as a threshold to prevent excessive orthogonality between sub-logits, with $\theta$= 0.5. The $\max(\cdot)$ term uses ReLU activation function as the loss.
\subsection{Overall Objective Function}
Our proposed HCD method combines cross-entropy loss, KL divergence, and a diversity constraint on sub-logit, enabling the student model to capture the teacher’s heterogeneous knowledge effectively and improve classification accuracy. The total HCD loss is defined as follows:
\begin{equation} 
\adjustbox{scale=0.98}{$
\begin{aligned}
\mathcal{L}_{HCD
}=\mathcal{L}_{CE
}+\mathcal{L}_{CE}^{sub}+ \lambda\mathcal{L}_{KL}+\beta \mathcal{L}_{KL}^{sub}+\omega \mathcal{L}_{orth
},
\end{aligned}
$}
\end{equation} 
where $\mathcal{L}_{KL}$ denotes the standard KL divergence. $\lambda$, $\beta$, and $\omega$ are the weight coefficients used to balance their contributions, and detailed algorithm in the \textit{Supplementary Material}.

\section{Experiment}
\subsection{Experiment settings}
\textbf{Datasets.} We evaluate the performance of the HCD method using the CIFAR-100~\cite{krizhevsky2009learning} and ImageNet-1K~\cite{deng2009imagenet} datasets. Specifically, CIFAR-100 consists of 100 classes, with a total of 50,000 samples, including 500 training images and 100 test images per class, each with a resolution of 32$\times$32 pixels. To ensure consistency with our model’s input requirements in heterogeneous distillation, we resize all CIFAR-100 images to 224$\times$224 pixels before training. In contrast, ImageNet-1K contains 1,000 classes, with approximately 1.28 million training images and 50,000 validation images, each with a resolution of 224$\times$224 pixels. CUB-200-2021~\cite{wah2011caltech} consists of 11,788 images from 200 bird species, widely used for fine-grained classification with annotations like bounding boxes and keypoints. FGVC-Aircraft~\cite{maji2013fine} consists of 10,000 images across 100 aircraft models, used for fine-grained classification tasks.\\
\textbf{Implementation details}, including comparative methods, model architectures, and training details, can be found in the supplementary material due to page constraints.
\begin{table*}[htbp]
  \centering
  \renewcommand{\arraystretch}{0.8}
  \resizebox{0.92\linewidth}{!}{%
    \begin{tabular}{cc|cc|ccccc|cccccccc}
      \tophline
      \multirow{2}{*}{Teacher} & \multirow{2}{*}{Student} & \multicolumn{2}{c|}{From Scratch} & \multicolumn{5}{c|}{Feature-based} & \multicolumn{8}{c}{Logit-based} \\
      \cmidrule(lr){3-17} 
                               &                          & T:Acc    & S:Acc    & FitNet & RKD   & CRD   & 
                               PAT   &
                               RSD& KD    & DKD   & DIST  & OFA
                               &TeKAP&   LDRLD&HCD &$\Delta$ \\
      \midhline
      Swin-T     & ResNet18     & 89.26 & 74.01 & 78.87 & 74.11 & 77.63&81.22 &\bf83.92 & 78.74 & 80.26 & 77.75 & 80.54&81.38&82.17 &\textbf{82.78}&\textbf{+4.04}\\
      ViT-S      & ResNet18     & 92.04 & 74.01 & 77.71 & 73.72 & 76.60 &80.11 &\bf81.50& 77.26& 78.10 & 76.49 &  80.15&79.06&80.36&\textbf{81.33} &\textbf{+4.07}\\
      Mixer-B/16 & ResNet18     & 87.29 & 74.01 & 77.15 & 73.75 & 76.42 & 80.07&\bf81.85&77.79 & 78.67 & 76.36 &  79.39 &80.05&80.69&\textbf{81.53}&\textbf{+4.24}\\
      Swin-T     & MobileNetV2     & 89.26 & 73.68 & 74.28 & 69.00 & 79.80 & 78.78&\bf83.68&74.68 & 71.07 & 72.89 &  80.98 &80.23&81.64&\textbf{82.19}&\textbf{+7.51}\\
      ViT-S      & MobileNetV2     & 92.04 & 73.68 & 73.54 & 68.46 & 78.14 & 78.87&\bf81.68&72.77 & 69.80 & 72.54 &  78.45&78.41 &79.21&\textbf{80.81}&\textbf{+8.04}\\
      Mixer-B/16 & MobileNetV2     & 87.29 & 73.68 & 73.78 & 68.95 & 78.15 &78.62&\bf81.74& 73.33 & 70.20 & 73.26 &  78.78& 79.89&80.64&\textbf{81.09}&\textbf{+7.76}\\
      \bottomline
    \end{tabular}
  }
    \caption{Evaluation of the top-1 accuracy (\%) of student using ViT-based heterogeneous models on the CIFAR100.}
    \label{tab:cifar}
\end{table*}

\begin{table*}
  \centering
  \small
 \renewcommand{\arraystretch}{0.8}
  \resizebox{0.92\linewidth}{!}{%
  \begin{tabular}{cc|cc|ccccc|ccccccc}
    \tophline
    \multirow{2}{*}{Teacher} & \multirow{2}{*}{Student} & \multicolumn{2}{c|}{From Scratch} & \multicolumn{5}{c|}{Feature-based} & \multicolumn{7}{c}{Logit-based}                                                                                                                                             \\
    \cmidrule(){3-16}
                             &                          & T:Acc                                & S:Acc                              & FitNet                           & CC    & RKD   & CRD & RSD & KD                & DKD                            & DIST                           & OFA  &TeKAP  & HCD   &$\Delta$                 \\
    \midhline
    Swin-T                   & ResNet18                 & 81.38                             & 69.75                           & 71.18                & 70.07 & 68.89 & 69.09&\bf72.13 & 71.14             & 71.10                          & 70.91                          & 71.85  &71.25  & \bf{71.91}      &\textbf{+0.77}          \\
    Mixer-B/16               & ResNet18                 & 76.62                             & 69.75                           & 70.78                            & 70.05 & 69.46 & 68.40 &\bf71.41& 70.89 & 69.89                          & 70.66                          & 71.38 &70.95 & \bf{71.66}  &\textbf{+0.77}                \\
    DeiT-T                   & MobileNetV2              & 72.17                             & 68.87                           & 70.95                            & 70.69 & 69.72 & 69.60 &\bf72.18& 70.87             & 70.14                          & 71.08              & \bf{71.39}        &70.23  &   71.38     &+0.51  \\
    Swin-T                   & MobileNetV2              & 81.38                             & 68.87                           & 71.75                            & 70.69 & 67.52 & 69.58 &\bf72.36& 72.05 & 71.71                          & 71.76                          & 72.32 &72.50 & \bf{72.72}   &\textbf{+0.67}               \\
    Mixer-B/16               & MobileNetV2              & 76.62                             & 68.87                           & 71.59                            & 70.79 & 69.86 & 68.89 &\bf71.90& 71.92 & 70.93                          & 71.74                          & 72.12  &72.01& \bf{72.32}  &\textbf{+0.40}               \\
    \bottomline
  \end{tabular}}
  \caption{Evaluation of the top-1 accuracy (\%) of student using ViT-based heterogeneous models on the ImageNet-1K.}
   \label{tab:imagenet-1k}
\end{table*}
\begin{table*}[!ht]
  \centering
\large
\renewcommand{\arraystretch}{1.9}
\scalebox{0.55}{
    \begin{tabular}{|c|c|c|c|c|c|c|c|c|c|c|c|c|c|c|c|c|c|c|c|}
    \tophline
    \tophline
    \multicolumn{20}{|c|}{ResNet34 (teacher): 73.31\% Top-1, 91.42\% Top-5 accuracy.
ResNet18 (student): 69.75\% Top-1, 89.07\% Top-5 accuracy.} \\
    \tophline
    \textbf{Features} & \multicolumn{1}{c|}{AT}& \multicolumn{1}{c|}{OFD} & \multicolumn{1}{c|}{CRD} & \multicolumn{1}{c|}{ReviewKD} & \multicolumn{1}{c|}{FCFD} & \multicolumn{1}{c|}{CAT-KD} &\multicolumn{1}{c|}{RSD}& \textbf{Logits} &\multicolumn{1}{c|}{KD} & \multicolumn{1}{c|}{CTKD} & \multicolumn{1}{c|}{DKD}  & \multicolumn{1}{c|}{LSKD} & \multicolumn{1}{c|}{SDD}& 
    \multicolumn{1}{c|}{WTTM}&\multicolumn{1}{c|}{TeKAP}& \multicolumn{1}{c|}{RLD}& \multicolumn{1}{c|}{LDRLD}& \multicolumn{1}{c|}{HCD}&\multicolumn{1}{c|}{$\Delta$}\\
    \midhline
    Top-1  & \multicolumn{1}{c|}{70.69}& \multicolumn{1}{c|}{70.81}& \multicolumn{1}{c|}{71.17} & \multicolumn{1}{c|}{71.61}  & \multicolumn{1}{c|}{\textbf{72.24}} &\multicolumn{1}{c|}{71.61} & \multicolumn{1}{c|}{72.18}& Top-1 & \multicolumn{1}{c|}{70.66} & \multicolumn{1}{c|}{71.32} & \multicolumn{1}{c|}{71.70} & \multicolumn{1}{c|}{71.42} & \multicolumn{1}
    {c|}{71.44} & \multicolumn{1}
    {c|}{\textbf{72.19}}& \multicolumn{1}
    {c|}{71.35}&\multicolumn{1}
    {c|}{71.91}&\multicolumn{1}
    {c|}{71.88}&\multicolumn{1}
    {c|}{72.18}&\multicolumn{1}
    {c|}{\textbf{+1.52}}\\
    \midhline
    Top-5  & \multicolumn{1}{c|}{90.01} & \multicolumn{1}{c|}{90.34}& \multicolumn{1}{c|}{90.13} & \multicolumn{1}{c|}{90.51} & \multicolumn{1}{c|}{\textbf{90.74}}  & \multicolumn{1}{c|}{90.45}&
    \multicolumn{1}{c|}{-} & Top-5 & \multicolumn{1}{c|}{89.88} & \multicolumn{1}{c|}{90.27} & \multicolumn{1}{c|}{90.31} & \multicolumn{1}{c|}{90.29} & \multicolumn{1}
    {c|}{90.05} & \multicolumn{1}{c|}{-} & \multicolumn{1}{c|}{90.54}& \multicolumn{1}{c|}{90.54}&\multicolumn{1}{c|}{90.59} &\multicolumn{1}{c|}{\textbf{90.64}}&\multicolumn{1}
    {c|}{\textbf{+0.76}}\\
    \bottomline
    \multicolumn{20}{|c|}{ResNet50 (teacher): 76.16\% Top-1, 92.87\% Top-5 accuracy.
MobileNetV1 (student): 68.87\% Top-1, 88.76\% Top-5 accuracy.} \\
    \tophline
    \textbf{Features} & \multicolumn{1}{c|}{AT}  & \multicolumn{1}{c|}{OFD}& \multicolumn{1}{c|}{CRD} & \multicolumn{1}{c|}{ReviewKD} & \multicolumn{1}{c|}{FCFD}  & \multicolumn{1}{c|}{CAT-KD}  & \multicolumn{1}{c|}{RSD}& \textbf{Logits} &      \multicolumn{1}{c|}{KD} & \multicolumn{1}{c|}{IPWD} & \multicolumn{1}{c|}{DKD}  & \multicolumn{1}{c|}{LSKD} & \multicolumn{1}{c|}{SDD}& 
    \multicolumn{1}{c|}{WTTM}& 
    \multicolumn{1}{c|}{TeKAP}& 
    \multicolumn{1}{c|}{RLD}&
    \multicolumn{1}{c|}{LDRLD}&
    \multicolumn{1}{c|}{HCD}&\multicolumn{1}{c|}{$\Delta$}\\
    \midhline
    Top-1  & \multicolumn{1}{c|}{70.18}  & \multicolumn{1}{c|}{71.25}& \multicolumn{1}{c|}{71.32} & \multicolumn{1}{c|}{72.56}  & \multicolumn{1}{c|}{\textbf{73.37}} & \multicolumn{1}{c|}{72.24} & \multicolumn{1}{c|}{73.08}& Top-1 & \multicolumn{1}{c|}{70.49} & \multicolumn{1}{c|}{72.65} & \multicolumn{1}{c|}{72.05} &  \multicolumn{1}{c|}{72.18} & \multicolumn{1}{c|}
    {72.24} & \multicolumn{1}{c|}{73.09}& \multicolumn{1}{c|}{72.87}& \multicolumn{1}{c|}{72.75}&\multicolumn{1}{c|}{73.12}&\multicolumn{1}{c|}{\textbf{73.23}}&\multicolumn{1}{c|}{\textbf{+2.74}}\\
    \midhline
    Top-5  & \multicolumn{1}{c|}{89.68} & \multicolumn{1}{c|}{90.34}& \multicolumn{1}{c|}{90.41} & \multicolumn{1}{c|}{91.00}  & \multicolumn{1}{c|}{\textbf{91.35}}  & \multicolumn{1}{c|}{91.13}
     & \multicolumn{1}{c|}{-}& Top-5 & \multicolumn{1}{c|}{89.92} & \multicolumn{1}{c|}{91.08} & \multicolumn{1}{c|}{91.05} & \multicolumn{1}{c|}{90.80} & \multicolumn{1}{c|}
    {90.71} & \multicolumn{1}{c|}{-}& \multicolumn{1}{c|}{91.05} & \multicolumn{1}{c|}{91.18}& \multicolumn{1}{c|}{91.43}&\multicolumn{1}{c|}{\textbf{91.46}}&\multicolumn{1}{c|}{\textbf{+1.54}}\\
    \bottomline
    \bottomline
    \end{tabular}%
    }
     \caption{Evaluation of the top-1 and top-5 accuracy (\%) of student using same-architecture on the ImageNet-1K.}
  \label{tab:imagenet}%
\end{table*}
\begin{table*}[!ht]
\begin{center}
\renewcommand{\arraystretch}{1.0}
\resizebox{0.92\linewidth}{!}{%
\begin{tabular}{c|c|cccccccc|cccccccc}
\tophline
        & \multicolumn{1}{c|}{Dataset} & \multicolumn{8}{c|}{CUB-200-2011}                                                                                                                     & \multicolumn{8}{c}{FGVC-Aircraft}                                                                                                                                                                            \\ 
        
    \tophline
        & \multicolumn{1}{c|}{Teacher's Acc} & \multicolumn{8}{c|}{Top-1,Top-5 for ViT-B: 82.76,  97.50; ViT-L: 85.28, 98.03}                                                                                                               & \multicolumn{8}{c}{Top-1,Top-5 for ViT-B: 74.56, 95.20; ViT-L: 88.48, 99.25}    \\    
        \cline{2-18}
 \multirow{1}{*}{Teacher} &  \multirow{2}{*}{Method} & \multicolumn{2}{c|}{VGG8}                             & \multicolumn{2}{c|}{ResNet20}                         & \multicolumn{2}{c|}{MobileNetV2}                      & \multicolumn{2}{c|}{ShuffleNetV2}   & \multicolumn{2}{c|}{VGG8}                             & \multicolumn{2}{c|}{ResNet20}                         & \multicolumn{2}{c|}{MobileNetV2}                      & \multicolumn{2}{c}{ShuffleNetV2}   \\ \cline{3-18}
      & \multicolumn{1}{c|}{}     & \multicolumn{1}{c|}{Top-1} & \multicolumn{1}{c|}{Top-5} & \multicolumn{1}{c|}{Top-1} & \multicolumn{1}{c|}{Top-5} & \multicolumn{1}{c|}{Top-1} & \multicolumn{1}{c|}{Top-5} & \multicolumn{1}{c|}{Top-1} & Top-5 & \multicolumn{1}{c|}{Top-1} & \multicolumn{1}{c|}{Top-5} & \multicolumn{1}{c|}{Top-1} & \multicolumn{1}{c|}{Top-5} & \multicolumn{1}{c|}{Top-1} & \multicolumn{1}{c|}{Top-5} & \multicolumn{1}{c|}{Top-1} & Top-5 \\ \hline
        & From Scratch                     & \multicolumn{1}{c|}{43.91}     & \multicolumn{1}{c|}{70.97}     & \multicolumn{1}{c|}{44.14}     & \multicolumn{1}{c|}{72.68}     & \multicolumn{1}{c|}{38.23}     & \multicolumn{1}{c|}{65.53}     & \multicolumn{1}{c|}{53.49}     &   \multicolumn{1}{c|}{78.34}   &  \multicolumn{1}{c|}{66.06}    & \multicolumn{1}{c|}{88.39}     & \multicolumn{1}{c|}{63.81}     & \multicolumn{1}{c|}{88.21}     & \multicolumn{1}{c|}{70.24}     & \multicolumn{1}{c|}{90.46}     & \multicolumn{1}{c|}{62.26}     &   \multicolumn{1}{c}{90.21}    \\
        \cline{2-18}
       
  & KD                           & \multicolumn{1}{c|}{56.08}     & \multicolumn{1}{c|}{83.81}     & \multicolumn{1}{c|}{42.04}     & \multicolumn{1}{c|}{73.70}    & \multicolumn{1}{c|}{44.13}      & \multicolumn{1}{c|}{57.66}  & \multicolumn{1}{c|}{42.77}     & \multicolumn{1}{c|}{72.94}        & \multicolumn{1}{c|}{70.75}     & \multicolumn{1}{c|}{93.31}     & \multicolumn{1}{c|}{56.04}     & \multicolumn{1}{c|}{87.40}     & \multicolumn{1}{c|}{66.64}     & \multicolumn{1}{c|}{91.39}     &   \multicolumn{1}{c|}{61.65}     & \multicolumn{1}{c}{89.80}      \\
   & DKD                          & \multicolumn{1}{c|}{61.06}     & \multicolumn{1}{c|}{86.52}     & \multicolumn{1}{c|}{38.21}     & \multicolumn{1}{c|}{70.19}     & \multicolumn{1}{c|}{53.33}     & \multicolumn{1}{c|}{80.95}     & \multicolumn{1}{c|}{49.78}     &   \multicolumn{1}{c|}{77.87}   & \multicolumn{1}{c|}{71.41}     & \multicolumn{1}{c|}{93.42}     & \multicolumn{1}{c|}{52.18}     & \multicolumn{1}{c|}{86.77}     & \multicolumn{1}{c|}{70.99}     & \multicolumn{1}{c|}{93.31}     & \multicolumn{1}{c|}{65.74}     &    \multicolumn{1}{c}{90.82}  \\ 
 SigLIP2       & WTTM                        & \multicolumn{1}{c|}{59.35}     & \multicolumn{1}{c|}{84.76}     & \multicolumn{1}{c|}{42.84}     & \multicolumn{1}{c|}{73.27}     & \multicolumn{1}{c|}{56.23}     & \multicolumn{1}{c|}{82.45}     & \multicolumn{1}{c|}{55.63}     &    \multicolumn{1}{c|}{81.53}   & \multicolumn{1}{c|}{52.60}     & \multicolumn{1}{c|}{87.22}     & \multicolumn{1}{c|}{60.81}     & \multicolumn{1}{c|}{88.98}     & \multicolumn{1}{c|}{76.12}     & \multicolumn{1}{c|}{\bf 94.84}     & \multicolumn{1}{c|}{8.00}     &     \multicolumn{1}{c}{20.98}  \\
 (ViT-B)       & RKKD                        & \multicolumn{1}{c|}{59.06}     & \multicolumn{1}{c|}{83.85}     & \multicolumn{1}{c|}{38.70}     & \multicolumn{1}{c|}{68.64}     & \multicolumn{1}{c|}{50.76}     & \multicolumn{1}{c|}{78.98}     & \multicolumn{1}{c|}{46.32}     &   \multicolumn{1}{c|}{74.09}   & \multicolumn{1}{c|}{73.30}     & \multicolumn{1}{c|}{93.24}     & \multicolumn{1}{c|}{62.35}     & \multicolumn{1}{c|}{88.96}     & \multicolumn{1}{c|}{72.88}     & \multicolumn{1}{c|}{92.77}     & \multicolumn{1}{c|}{70.18}     & \multicolumn{1}{c}{91.78}     \\
        & LDRLD                       & \multicolumn{1}{c|}{\bf 63.34}     & \multicolumn{1}{c|}{\bf 87.00}     & \multicolumn{1}{c|}{50.60}     & \multicolumn{1}{c|}{77.60}     & \multicolumn{1}{c|}{\bf 59.15}     & \multicolumn{1}{c|}{\bf 84.07}     & \multicolumn{1}{c|}{\bf 55.80}     &     \multicolumn{1}{c|}{\bf 81.79}  & \multicolumn{1}{c|}{70.30}     & \multicolumn{1}{c|}{91.33}     & \multicolumn{1}{c|}{59.08}     & \multicolumn{1}{c|}{86.44}     & \multicolumn{1}{c|}{75.28}     & \multicolumn{1}{c|}{94.18}     & \multicolumn{1}{c|}{71.18}     &  \multicolumn{1}{c}{92.17}    \\
        & HCD (Ours)                          & \multicolumn{1}{c|}{62.16}     & \multicolumn{1}{c|}{86.23}     & \multicolumn{1}{c|}{\bf 50.98}     & \multicolumn{1}{c|}{\bf 78.89}     & \multicolumn{1}{c|}{ 55.32}     & \multicolumn{1}{c|}{81.25}     & \multicolumn{1}{c|}{54.70}     &   \multicolumn{1}{c|}{80.74}   & \multicolumn{1}{c|}{\bf 75.04}     & \multicolumn{1}{c|}{\bf 92.53}     & \multicolumn{1}{c|}{\bf 66.17}     & \multicolumn{1}{c|}{\bf 90.11}     & \multicolumn{1}{c|}{\bf 76.23}     & \multicolumn{1}{c|}{ 93.97}     & \multicolumn{1}{c|}{\bf 72.76}     &    \multicolumn{1}{c}{\bf 92.33}   \\ 
        \hline
        & From Scratch                     & \multicolumn{1}{c|}{43.91}     & \multicolumn{1}{c|}{70.97}     & \multicolumn{1}{c|}{44.14}     & \multicolumn{1}{c|}{72.68}     & \multicolumn{1}{c|}{38.23}     & \multicolumn{1}{c|}{65.53}     & \multicolumn{1}{c|}{53.49}     &   \multicolumn{1}{c|}{78.34}   &  \multicolumn{1}{c|}{66.06}    & \multicolumn{1}{c|}{88.39}     & \multicolumn{1}{c|}{63.81}     & \multicolumn{1}{c|}{88.21}     & \multicolumn{1}{c|}{70.24}     & \multicolumn{1}{c|}{90.46}     & \multicolumn{1}{c|}{62.26}     &   \multicolumn{1}{c}{90.21}    \\
        \cline{2-18}
       %\cdashline{2-18}[2pt/2pt]
 & KD                           & \multicolumn{1}{c|}{56.32}     & \multicolumn{1}{c|}{84.02}     & \multicolumn{1}{c|}{43.18}     & \multicolumn{1}{c|}{74.58}     & \multicolumn{1}{c|}{44.24}     & \multicolumn{1}{c|}{74.09}     & \multicolumn{1}{c|}{44.25}     &   \multicolumn{1}{c|}{73.51}   & \multicolumn{1}{c|}{75.66}     & \multicolumn{1}{c|}{94.54}     & \multicolumn{1}{c|}{62.50}     & \multicolumn{1}{c|}{90.19}     & \multicolumn{1}{c|}{73.66}     & \multicolumn{1}{c|}{93.76}     & \multicolumn{1}{c|}{72.00}     &   \multicolumn{1}{c}{92.47}    \\
  & DKD                          & \multicolumn{1}{c|}{58.31}     & \multicolumn{1}{c|}{84.82}     & \multicolumn{1}{c|}{35.23}     & \multicolumn{1}{c|}{67.42}     & \multicolumn{1}{c|}{52.19}     & \multicolumn{1}{c|}{80.86}     & \multicolumn{1}{c|}{49.55}     &  \multicolumn{1}{c|}{77.30}        & \multicolumn{1}{c|}{75.55}     & \multicolumn{1}{c|}{\bf 94.59}     & \multicolumn{1}{c|}{53.53}     & \multicolumn{1}{c|}{87.46}     & \multicolumn{1}{c|}{75.28}     & \multicolumn{1}{c|}{94.18}     & \multicolumn{1}{c|}{72.04}     &     \multicolumn{1}{c}{93.49} \\
 SigLIP2        & WTTM              & \multicolumn{1}{c|}{60.17}     & \multicolumn{1}{c|}{85.38}     & \multicolumn{1}{c|}{40.94}     & \multicolumn{1}{c|}{70.61}        & \multicolumn{1}{c|}{55.63}     & \multicolumn{1}{c|}{81.48}     & \multicolumn{1}{c|}{54.09}         & \multicolumn{1}{c|}{79.89}     & \multicolumn{1}{c|}{62.14}     & \multicolumn{1}{c|}{90.18}     & \multicolumn{1}{c|}{65.38}     & \multicolumn{1}{c|}{89.86}   &  \multicolumn{1}{c|}{78.37} & \multicolumn{1}{c|}{94.95}     & \multicolumn{1}{c|}{70.42}     &   \multicolumn{1}{c}{90.22}   \\ 
 (ViT-L)        & RKKD                         & \multicolumn{1}{c|}{56.53}     & \multicolumn{1}{c|}{82.72}     & \multicolumn{1}{c|}{36.95}     & \multicolumn{1}{c|}{67.74}     & \multicolumn{1}{c|}{49.83}     & \multicolumn{1}{c|}{77.70}     & \multicolumn{1}{c|}{47.17}     &   \multicolumn{1}{c|}{74.28} &\multicolumn{1}{c|}{76.57}      & \multicolumn{1}{c|}{94.54}     & \multicolumn{1}{c|}{66.16}     & \multicolumn{1}{c|}{90.82}     & \multicolumn{1}{c|}{76.12}     & \multicolumn{1}{c|}{94.30}     & \multicolumn{1}{c|}{75.64}     &  \multicolumn{1}{c}{93.43}    \\
        & LDRLD                       & \multicolumn{1}{c|}{62.55}     & \multicolumn{1}{c|}{86.28}     & \multicolumn{1}{c|}{49.23}     & \multicolumn{1}{c|}{76.32}     & \multicolumn{1}{c|}{\bf 58.75}     & \multicolumn{1}{c|}{\bf 85.89}     & \multicolumn{1}{c|}{\bf 56.20}     &     \multicolumn{1}{c|}{\bf 81.91}  & \multicolumn{1}{c|}{74.89}     & \multicolumn{1}{c|}{93.73}     & \multicolumn{1}{c|}{58.85}     & \multicolumn{1}{c|}{86.01}     & \multicolumn{1}{c|}{78.76}     & \multicolumn{1}{c|}{94.69}     & \multicolumn{1}{c|}{77.05}     &     \multicolumn{1}{c}{94.00} \\ 
         & HCD (Ours)                         & \multicolumn{1}{c|}{\bf 63.21}     & \multicolumn{1}{c|}{\bf86.90}     & \multicolumn{1}{c|}{\bf 50.04}     & \multicolumn{1}{c|}{\bf 77.34}     & \multicolumn{1}{c|}{ 56.14}     &   \multicolumn{1}{c|}{82.38}  &  \multicolumn{1}{c|}{ 54.36}   &   \multicolumn{1}{c|}{ 80.50}     & \multicolumn{1}{c|}{\bf 77.08}     & \multicolumn{1}{c|}{ 93.34}     & \multicolumn{1}{c|}{\bf 68.47}     & \multicolumn{1}{c|}{\bf 90.82}     & \multicolumn{1}{c|}{\bf 80.02}     & \multicolumn{1}{c|}{\bf 95.53}     & \multicolumn{1}{c|}{\bf 77.65}     &      \multicolumn{1}{c}{\bf 94.23} \\\bottomline
\end{tabular}
}
\caption{Image classification accuracy (\%) on fine-grained datasets using SigLIP2 as a teacher and CNN as a student.}
\label{siglip2}
\end{center}
\end{table*}
\begin{table*}[!ht]
\begin{center}
\renewcommand{\arraystretch}{1.0}
\resizebox{0.92\linewidth}{!}{%
\begin{tabular}{c|c|cccccccc|cccccccc}
\tophline
        & \multicolumn{1}{c|}{Dataset} & \multicolumn{8}{c|}{CUB-200-2011}                                                                                                                     & \multicolumn{8}{c}{FGVC-Aircraft}                                                                                                                                                                           \\ 
            \tophline
        & \multicolumn{1}{c|}{Teacher's Acc} & \multicolumn{8}{c|}{Top-1,Top-5 for ViT-S: 84.35, 96.29; ViT-L: 88.99, 98.12}                                                                                                                     & \multicolumn{8}{c}{Top-1,Top-5 for ViT-S: 71.68, 94.96;  ViT-L: 83.41, 97.06}  \\    
        \cline{2-18}
 \multirow{1}{*}{Teacher} & \multirow{2}{*}{Method} & \multicolumn{2}{c|}{VGG8}                             & \multicolumn{2}{c|}{ResNet20}                         & \multicolumn{2}{c|}{MobileNetV2}                      & \multicolumn{2}{c|}{ShuffleNetV2}   & \multicolumn{2}{c|}{VGG8}                             & \multicolumn{2}{c|}{ResNet20}                         & \multicolumn{2}{c|}{MobileNetV2}                      & \multicolumn{2}{c}{ShuffleNetV2}   \\  \cline{3-18}
        & \multicolumn{1}{c|}{}     & \multicolumn{1}{c|}{Top-1} & \multicolumn{1}{c|}{Top-5} & \multicolumn{1}{c|}{Top-1} & \multicolumn{1}{c|}{Top-5} & \multicolumn{1}{c|}{Top-1} & \multicolumn{1}{c|}{Top-5} & \multicolumn{1}{c|}{Top-1} & Top-5 & \multicolumn{1}{c|}{Top-1} & \multicolumn{1}{c|}{Top-5} & \multicolumn{1}{c|}{Top-1} & \multicolumn{1}{c|}{Top-5} & \multicolumn{1}{c|}{Top-1} & \multicolumn{1}{c|}{Top-5} & \multicolumn{1}{c|}{Top-1} & Top-5 \\ \hline
        & From Scratch                     & \multicolumn{1}{c|}{46.50}     & \multicolumn{1}{c|}{72.76}     & \multicolumn{1}{c|}{50.50}     & \multicolumn{1}{c|}{76.77}     & \multicolumn{1}{c|}{50.38}     & \multicolumn{1}{c|}{76.30}     & \multicolumn{1}{c|}{53.49}     &   \multicolumn{1}{c|}{78.34}   &  \multicolumn{1}{c|}{68.20}    & \multicolumn{1}{c|}{88.66}     & \multicolumn{1}{c|}{62.19}     & \multicolumn{1}{c|}{87.57}     & \multicolumn{1}{c|}{69.19}     & \multicolumn{1}{c|}{90.25}     & \multicolumn{1}{c|}{72.61}     &   \multicolumn{1}{c}{91.75}    \\
        \cline{2-18}
       
  & KD                           & \multicolumn{1}{c|}{60.48}     & \multicolumn{1}{c|}{84.98}     & \multicolumn{1}{c|}{50.78}     & \multicolumn{1}{c|}{77.55}    & \multicolumn{1}{c|}{60.75}      & \multicolumn{1}{c|}{84.17}  & \multicolumn{1}{c|}{60.22}     & \multicolumn{1}{c|}{83.22}        & \multicolumn{1}{c|}{67.75}     & \multicolumn{1}{c|}{90.82}     & \multicolumn{1}{c|}{59.74}     & \multicolumn{1}{c|}{86.80}     & \multicolumn{1}{c|}{69.88}     & \multicolumn{1}{c|}{92.14}     &   \multicolumn{1}{c|}{70.72}     & \multicolumn{1}{c}{90.34}      \\
   & DKD                          & \multicolumn{1}{c|}{64.82}     & \multicolumn{1}{c|}{87.02}     & \multicolumn{1}{c|}{46.48}     & \multicolumn{1}{c|}{77.34}     & \multicolumn{1}{c|}{64.03}     & \multicolumn{1}{c|}{86.78}     & \multicolumn{1}{c|}{62.63}     &   \multicolumn{1}{c|}{85.61}   & \multicolumn{1}{c|}{55.03}     & \multicolumn{1}{c|}{87.91}     & \multicolumn{1}{c|}{24.21}     & \multicolumn{1}{c|}{65.96}     & \multicolumn{1}{c|}{52.18}     & \multicolumn{1}{c|}{87.43}     & \multicolumn{1}{c|}{47.46}     &    \multicolumn{1}{c}{84.91}  \\ 
 DINOv2       & WTTM                        & \multicolumn{1}{c|}{\bf 70.71}     & \multicolumn{1}{c|}{\bf 89.21}     & \multicolumn{1}{c|}{52.66}     & \multicolumn{1}{c|}{78.37}     & \multicolumn{1}{c|}{61.06}     & \multicolumn{1}{c|}{84.16}     & \multicolumn{1}{c|}{56.32}     &    \multicolumn{1}{c|}{80.62}   & \multicolumn{1}{c|}{ 70.78}     & \multicolumn{1}{c|}{91.78}     & \multicolumn{1}{c|}{63.19}     & \multicolumn{1}{c|}{87.46}     & \multicolumn{1}{c|}{72.52}     & \multicolumn{1}{c|}{93.03}     & \multicolumn{1}{c|}{69.34}     &     \multicolumn{1}{c}{90.49}  \\
 (ViT-S)       & RKKD                        & \multicolumn{1}{c|}{61.41}     & \multicolumn{1}{c|}{85.17}     & \multicolumn{1}{c|}{51.14}     & \multicolumn{1}{c|}{77.22}     & \multicolumn{1}{c|}{59.77}     & \multicolumn{1}{c|}{83.34}     & \multicolumn{1}{c|}{60.77}     &   \multicolumn{1}{c|}{83.78}   & \multicolumn{1}{c|}{67.33}     & \multicolumn{1}{c|}{90.52}     & \multicolumn{1}{c|}{58.54}     & \multicolumn{1}{c|}{86.26}     & \multicolumn{1}{c|}{ 70.51}     & \multicolumn{1}{c|}{92.02}     & \multicolumn{1}{c|}{68.71}     & \multicolumn{1}{c}{90.34}     \\
        & LDRLD                       & \multicolumn{1}{c|}{68.43}     & \multicolumn{1}{c|}{87.97}     & \multicolumn{1}{c|}{56.66}     & \multicolumn{1}{c|}{82.38}     & \multicolumn{1}{c|}{\bf 71.30}     & \multicolumn{1}{c|}{\bf 90.70}     & \multicolumn{1}{c|}{\bf 68.81}     &     \multicolumn{1}{c|}{\bf 88.40}  & \multicolumn{1}{c|}{68.89}     & \multicolumn{1}{c|}{91.60}     & \multicolumn{1}{c|}{50.77}     & \multicolumn{1}{c|}{83.98}     & \multicolumn{1}{c|}{70.12}     & \multicolumn{1}{c|}{93.01}     & \multicolumn{1}{c|}{66.28}     &  \multicolumn{1}{c}{90.49}    \\
        & HCD  (Ours)                         & \multicolumn{1}{c|}{69.16}     & \multicolumn{1}{c|}{88.23}     & \multicolumn{1}{c|}{\bf 57.58}     & \multicolumn{1}{c|}{\bf 82.78}     & \multicolumn{1}{c|}{ 67.12}     & \multicolumn{1}{c|}{87.25}     & \multicolumn{1}{c|}{64.48}     &   \multicolumn{1}{c|}{86.64}   & \multicolumn{1}{c|}{\bf 74.20}     & \multicolumn{1}{c|}{\bf 92.17}     & \multicolumn{1}{c|}{\bf 65.17}     & \multicolumn{1}{c|}{\bf 88.57}     & \multicolumn{1}{c|}{\bf 78.49}     & \multicolumn{1}{c|}{\bf 94.66}     & \multicolumn{1}{c|}{\bf 75.76}     &    \multicolumn{1}{c}{\bf 92.56}   \\ 
        \hline
        & From Scratch                     & \multicolumn{1}{c|}{46.50}     & \multicolumn{1}{c|}{72.76}     & \multicolumn{1}{c|}{50.50}     & \multicolumn{1}{c|}{76.77}     & \multicolumn{1}{c|}{50.38}     & \multicolumn{1}{c|}{76.30}     & \multicolumn{1}{c|}{53.49}     &   \multicolumn{1}{c|}{78.34}   &  \multicolumn{1}{c|}{68.20}    & \multicolumn{1}{c|}{88.66}     & \multicolumn{1}{c|}{62.19}     & \multicolumn{1}{c|}{87.57}     & \multicolumn{1}{c|}{69.19}     & \multicolumn{1}{c|}{90.25}     & \multicolumn{1}{c|}{72.61}     &   \multicolumn{1}{c}{91.75}    \\
        \cline{2-18}
       % \cdashline{2-18}[2pt/2pt]
 & KD                           & \multicolumn{1}{c|}{60.94}     & \multicolumn{1}{c|}{83.95}     & \multicolumn{1}{c|}{50.69}     & \multicolumn{1}{c|}{76.65}     & \multicolumn{1}{c|}{57.75}     & \multicolumn{1}{c|}{81.83}     & \multicolumn{1}{c|}{62.16}     &   \multicolumn{1}{c|}{84.47}   & \multicolumn{1}{c|}{72.94}     & \multicolumn{1}{c|}{92.53}     & \multicolumn{1}{c|}{59.50}     & \multicolumn{1}{c|}{85.51}     & \multicolumn{1}{c|}{76.36}     & \multicolumn{1}{c|}{93.67}     & \multicolumn{1}{c|}{73.71}     &   \multicolumn{1}{c}{92.05}    \\
  & DKD                          & \multicolumn{1}{c|}{66.31}     & \multicolumn{1}{c|}{86.90}     & \multicolumn{1}{c|}{39.70}     & \multicolumn{1}{c|}{71.21}     & \multicolumn{1}{c|}{62.80}     & \multicolumn{1}{c|}{85.28}     & \multicolumn{1}{c|}{63.65}     &  \multicolumn{1}{c|}{85.12}        & \multicolumn{1}{c|}{61.33}     & \multicolumn{1}{c|}{90.19}     & \multicolumn{1}{c|}{27.90}     & \multicolumn{1}{c|}{71.86}     & \multicolumn{1}{c|}{53.41}     & \multicolumn{1}{c|}{88.48}     & \multicolumn{1}{c|}{49.95}     &     \multicolumn{1}{c}{85.45} \\
 DINOv2        & WTTM              & \multicolumn{1}{c|}{69.24}     & \multicolumn{1}{c|}{\bf 88.92}     & \multicolumn{1}{c|}{52.73}     & \multicolumn{1}{c|}{79.15}        & \multicolumn{1}{c|}{60.75}     & \multicolumn{1}{c|}{82.71}     & \multicolumn{1}{c|}{56.20}         & \multicolumn{1}{c|}{81.36}     & \multicolumn{1}{c|}{73.78}     & \multicolumn{1}{c|}{92.56}     & \multicolumn{1}{c|}{63.25}     & \multicolumn{1}{c|}{87.43}   &  \multicolumn{1}{c|}{76.18} & \multicolumn{1}{c|}{93.10}     & \multicolumn{1}{c|}{70.42}     &   \multicolumn{1}{c}{90.22}   \\ 
 (ViT-L)        & RKKD                         & \multicolumn{1}{c|}{60.99}     & \multicolumn{1}{c|}{84.88}     & \multicolumn{1}{c|}{51.61}     & \multicolumn{1}{c|}{78.31}     & \multicolumn{1}{c|}{60.39}     & \multicolumn{1}{c|}{83.79}     & \multicolumn{1}{c|}{60.65}     &   \multicolumn{1}{c|}{83.24} &\multicolumn{1}{c|}{72.22}      & \multicolumn{1}{c|}{92.13}     & \multicolumn{1}{c|}{62.77}     & \multicolumn{1}{c|}{87.61}     & \multicolumn{1}{c|}{76.45}     & \multicolumn{1}{c|}{93.46}     & \multicolumn{1}{c|}{73.21}     &  \multicolumn{1}{c}{91.60}    \\
        & LDRLD                       & \multicolumn{1}{c|}{\bf 69.38}     & \multicolumn{1}{c|}{85.89}     & \multicolumn{1}{c|}{52.78}     & \multicolumn{1}{c|}{78.15}     & \multicolumn{1}{c|}{66.21}     & \multicolumn{1}{c|}{85.89}     & \multicolumn{1}{c|}{64.62}     &     \multicolumn{1}{c|}{84.85}  & \multicolumn{1}{c|}{72.52}     & \multicolumn{1}{c|}{92.44}     & \multicolumn{1}{c|}{56.47}     & \multicolumn{1}{c|}{85.30}     & \multicolumn{1}{c|}{75.31}     & \multicolumn{1}{c|}{93.37}     & \multicolumn{1}{c|}{70.57}     &     \multicolumn{1}{c}{91.10} \\ 
         & HCD (Ours)                        & \multicolumn{1}{c|}{68.48}     & \multicolumn{1}{c|}{86.80}     & \multicolumn{1}{c|}{\bf 57.04}     & \multicolumn{1}{c|}{\bf 81.34}     & \multicolumn{1}{c|}{\bf 67.14}     &   \multicolumn{1}{c|}{\bf 88.38}  &  \multicolumn{1}{c|}{\bf 65.65}   &   \multicolumn{1}{c|}{\bf 87.16}     & \multicolumn{1}{c|}{\bf 74.04}     & \multicolumn{1}{c|}{\bf 92.86}     & \multicolumn{1}{c|}{\bf 63.41}     & \multicolumn{1}{c|}{\bf 88.51}     & \multicolumn{1}{c|}{\bf 78.73}     & \multicolumn{1}{c|}{\bf 94.09}     & \multicolumn{1}{c|}{\bf 74.50}     &      \multicolumn{1}{c}{\bf 92.83} \\\bottomline
\end{tabular}
}
\caption{Image classification accuracy (\%) on fine-grained datasets using DINOv2 as a teacher and CNN as a student.}
\label{dinov2}
\end{center}
\end{table*}
\subsection{Main Results and Analysis} 
\textbf{Image Classification Results on CIFAR-100.} To evaluate the effectiveness of our proposed HCD method, we first conducted heterogeneous KD experiments on the CIFAR-100 dataset. Our approach, based on logit distillation, employs various characteristics of both the teacher and student. Teacher models include Vision Transformer-based architectures (e.g., Swin-T, ViT-S, Mixer-B/16) and student models consist of ResNet18, MobileNetV2. As shown in Table~\ref{tab:cifar}, HCD outperforms existing logit-based KD methods, including OFA, DKD, TeKAP, and LDRLD, achieving a significant accuracy improvement of 4-8\% over traditional knowledge distillation (Vanilla KD), demonstrating its superiority in heterogeneous distillation tasks. This effectiveness stems from the ability of HCD to leverage complementary information from heterogeneous architectures by mapping intermediate student features and deep teacher features to shared logits space, followed by sub-logit decomposition and orthogonal constraints to reduce semantic discrepancies, thereby effectively enhancing knowledge transfer.\\
\textbf{Image Classification Results on ImageNet-1K.} To evaluate the effectiveness of our proposed HCD method, we conducted heterogeneous and homogeneous KD experiments on the ImageNet-1K dataset using Top-1 and Top-1\&5 accuracy metrics, respectively. The homogeneous architectures used in our experiments include the ResNet and MobileNet series. As shown in Table~\ref{tab:imagenet-1k}, our method outperforms existing logit-based KD approaches, such as OFA, with a Top-1 accuracy improvement of 0.40\% to 0.77\% over traditional KD. We further expand experiments on homogeneous distillation, Table~\ref{tab:imagenet} demonstrates that our method achieves a Top-1 accuracy improvement of 1.52\% and 2.74\%, and a Top-5 accuracy improvement of 0.76\% and 1.54\% for ResNet18 and MobileNetV1, further confirming its superiority. This performance stems from leveraging complementary information (e.g., low-level texture information and high-level discriminative features) through feature mapping and sub-logit decomposition, exhibiting strong scalability across heterogeneous and homogeneous architectures.\\
\textbf{Image Classification Results on Fine-grained datasets.} To further evaluate the performance of HCD and assess its generalizability, we also extended the experiments to fine-grained datasets. Leveraging recent advancements in the powerful feature representation capabilities of vision foundation models (VFMs), we employed the visual encoders of DINOv2~\cite{oquab2024dinov} and SigLIP2~\cite{tschannen2025siglip} as teacher models, with various CNNs acting as student models. As shown in Tables~\ref{siglip2} and \ref{dinov2}, compared to exsiting KD methods (e.g., DKD,LDRLD), HCD achieves superior performance on the FGVC-Aircraft. This results demonstrate that HCD effectively uses heterogeneous complementary information to enhance feature representation in transfer learning for VFMs. However, the FGVC-Aircraft images exhibit distinct category differences, while CUB-200 presents greater learning difficulty due to its fine-grained features, resulting in better performance in heterogeneous distillation for the former in Tables~\ref{siglip2} and \ref{dinov2}. However, HCD outperforms LDRLD slightly on the CUB200 dataset, likely due to the fine-grained nature of the task, where LDRLD excels in capturing inter-class relationships.
\subsection{Ablation Studies}
\textbf{Impact of the number of sub-logits ($n$).} We conducted an ablation study to investigate the impact of the number of decoupled sub-logits on the student's performance, using Swin-Tiny as the teacher and ResNet18 as the student, with $n$ ranging from 1 to 8. Table~\ref{ablation of n} indicates that $n$=4 yields the optimal performance, as four sub-logits strike a balance between student's performance and computational cost, optimizes diversity and capacity while enhancing heterogeneous knowledge transfer. Experimental analysis further revealed that at $n$=1, sub-logits compressed teacher’s global features and the student’s local patterns, leading to insufficient knowledge transfer, increased learning difficulty, and reducing generalization. However, at $n$=8, excessive sub-logits introduced redundant noise and computational overhead, resulting in performance degradation.
\begin{table}[!ht]
\centering
\scalebox{0.83}{
\begin{tabular}{c|c|c|c|c|c|c}
\tophline
T:Swin-Tiny & 89.26 & $n$ = 1 & $n$ = 2 & $n$ = 4 & $n$ = 6 & $n$ = 8 \\
\cline{2-7}
S:ResNet18 & 74.01 & 81.53& 81.84 & \bf{82.78} & 82.66 & 82.60 \\
\midhline
Time (s/epoch) & - & \bf 13.66& 13.78 & 14.32 & 14.56 & 14.82 \\
\bottomline
\end{tabular}}
\caption{Ablation study of $n$ on the CIFAR-100 dateset with 8 NVIDIA RTX 3090 GPUs.}
\label{ablation of n}
\end{table}\\
\textbf{Impact of different loss functions.} We kept the weight coefficients of the cross-entropy loss ($\mathcal{L}_{CE}^{sub}$ and $\mathcal{L}_{CE}$) constant and conducted ablation experiments to evaluate the impact of different loss functions on the CIFAR-100 dataset, as shown in Table~\ref{Impact of the loss}. Ablating the loss components reveals that each individual loss function yields a significant performance improvement over the baseline (e.g., over 3.0\% increase in Top-1 accuracy). Combining the three losses ($\mathcal{L}_{KD}$, $\mathcal{L}_{KL}^{sub}$, and $\mathcal{L}_{orth}$) yields the best performance, highlighting the effectiveness of our multi-loss optimization.
\begin{table}[!ht]
\centering
\resizebox{\linewidth}{!}{
\begin{tabular}{ccc|cccccc}
\tophline
\multirow{2}*{$\mathcal{L}_{KD}$} &\multirow{2}*{$\mathcal{L}_{KL}^{sub}$}& \multirow{2}*{$\mathcal{L}_{orth}$}&Swin-T& Mixer-B/16 & Swin-T  & ViT-S \\
&  &  & ResNet18& MobileNetV2 & MobileNetV2 & MobileNetV2 \\
\midhline
- & - & - & 74.01 & 73.68 & 73.68 & 73.68 \\
\checkmark & &  &  78.74  & 73.33 & 74.68  & 72.77  \\
\checkmark & \checkmark&  & 82.32 (+3.58) & 80.52 (+7.19) & 81.42 (+6.74)  & 80.39 (+7.62) \\
\checkmark & \checkmark & \checkmark & \textbf{82.78} (+4.04) & \textbf{81.09} (+7.76) & \textbf{82.19} (+7.51) & \textbf{80.81} (+8.04) \\
\bottomline
\end{tabular}
    }
\caption{Ablation study of different losses on student’s performance on the CIFAR100 dataset.}
\label{Impact of the loss}
\end{table}

\begin{comment}
\begin{table}[!ht]
\vspace{-8pt}
\centering
\resizebox{\linewidth}{!}{
\begin{tabular}{cccc|ccccc}
\tophline
\multirow{2}*{Stage:1} &\multirow{2}*{Stage:2}& \multirow{2}*{Stage:3}&\multirow{2}*{Stage:4}& Swin-T &  ViT-S &Swin-T& ViT-S \\
&  &  & &ResNet18& ResNet18 & MobileNetV2 & MobileNetV2 \\
\midhline
- & - & - & - &74.01& 74.01 & 73.68 & 73.68 \\
\checkmark & &  & &81.82 (\textcolor{deepgreen}{+3.75}) & 68.95 (\textcolor{deepgreen}{+4.35}) & 69.23 (\textcolor{deepgreen}{+4.63})  & 76.92 (\textcolor{deepgreen}{+5.10}) \\
\checkmark & \checkmark&  &  &82.34 (\textcolor{deepgreen}{+3.75}) & 68.95 (\textcolor{deepgreen}{+4.35}) & 69.23 (\textcolor{deepgreen}{+4.63})  & 76.92 (\textcolor{deepgreen}{+5.10}) \\
\checkmark & \checkmark & \checkmark &  & 82.46 (\textcolor{deepgreen}{+3.96}) & 69.27 (\textcolor{deepgreen}{+4.67}) &69.52 (\textcolor{deepgreen}{+4.92}) & 77.07 (\textcolor{deepgreen}{+5.25}) \\
\checkmark & \checkmark & \checkmark &\checkmark & \textbf{82.91} (\textcolor{deepgreen}{+3.96}) & \textbf{81.06} (\textcolor{deepgreen}{+4.67}) & \textbf{69.52} (\textcolor{deepgreen}{+4.92}) & \textbf{77.07} (\textcolor{deepgreen}{+5.25}) \\
\bottomline
\end{tabular}}
\vspace{-0.3cm}
\caption{Ablation Study of the number of stages on student’s performance with HCD on CIFAR100 dataset.}
\label{Impact of the stages}
\vspace{-22pt}
\end{table}
\end{comment}

\section{Conclusion}
\label{sec:conclusion}
In this work, we propose Heterogeneous Complementary Distillation (HCD), a novel framework that leverages complementary teacher-student feature representations through the Complementary Feature Mapper module to align them in shared logits space. We also propose Sub-logit Decoupled Distillation that decomposes these logits into $n$ sub-logits, fused with teacher logits for specialized knowledge transfer, while an orthogonality loss ensures sub-logit diversity and non-redundancy knowledge. Extensive experiments on CIFAR-100, ImageNet-1K, and Fine-grained datasets demonstrate that HCD outperforms state-of-the-art KD methods, enhancing student robustness and generalization by effectively integrating heterogeneous knowledge while preserving student-specific strengths.

\section{Acknowledgments}
The project was supported in part by Guangdong Major Project of Basic and Applied Basic Research (2023B0303000016), in part by National Natural Science Foundation of China (U21A20487, 62273082), in part by Shenzhen Technology Project (GJHZ20240218112504008), in part by Guangdong Technology Project (2023TX07Z126), in part by Shenzhen High-tech Zone Development Special Plan Innovation Platform Construction Project, the proof of concept center for high precision and high resolution 4D imaging.

\bibliography{aaai2026}

@inproceedings{he2016deep,
  title={Deep residual learning for image recognition},
  author={He, Kaiming and Zhang, Xiangyu and Ren, Shaoqing and Sun, Jian},
  booktitle={CVPR},
  pages={770--778},
  year={2016}
}

@inproceedings{yang2022cross,
  title={Cross-image relational knowledge distillation for semantic segmentation},
  author={Yang, Chuanguang and Zhou, Helong and An, Zhulin and Jiang, Xue and Xu, Yongjun and Zhang, Qian},
  booktitle={CVPR},
  pages={12319--12328},
  year={2022}
}

@inproceedings{zhao2024detrs,
  title={Detrs beat yolos on real-time object detection},
  author={Zhao, Yian and Lv, Wenyu and Xu, Shangliang and Wei, Jinman and Wang, Guanzhong and Dang, Qingqing and Liu, Yi and Chen, Jie},
  booktitle={CVPR},
  pages={16965--16974},
  year={2024}
}

@inproceedings{liang2023open,
  title={Open-vocabulary semantic segmentation with mask-adapted clip},
  author={Liang, Feng and Wu, Bichen and Dai, Xiaoliang and Li, Kunpeng and Zhao, Yinan and Zhang, Hang and Zhang, Peizhao and Vajda, Peter and Marculescu, Diana},
  booktitle={CVPR},
  pages={7061--7070},
  year={2023}
}

@inproceedings{sandler2018mobilenetv2,
  title={Mobilenetv2: Inverted residuals and linear bottlenecks},
  author={Sandler, Mark and Howard, Andrew and Zhu, Menglong and Zhmoginov, Andrey and Chen, Liang-Chieh},
  booktitle={CVPR},
  pages={4510--4520},
  year={2018}
}

@inproceedings{hinton2015distilling,
  title={Distilling the knowledge in a neural network},
  author={Hinton, Geoffrey and Vinyals, Oriol and Dean, Jeff and others},
  booktitle={NeurIPS Workshop},
  year={2014},
}

@article{kullback1951information,
  title={On information and sufficiency},
  author={Kullback, Solomon and Leibler, Richard A},
  journal={The Annals of Mathematical Statistics},
  volume={22},
  number={1},
  pages={79--86},
  year={1951},
  publisher={JSTOR}
}

@inproceedings{zhao2022decoupled,
  title={Decoupled Knowledge Distillation},
  author={Zhao, Borui and Cui, Quan and Song, Renjie and Qiu, Yiyu and Liang, Jiajun},
  booktitle={CVPR},
  pages={11953--11962},
  year={2022}
}

@inproceedings{yang2023knowledge,
  title={From knowledge distillation to self-knowledge distillation: A unified approach with normalized loss and customized soft labels},
  author={Yang, Zhendong and Zeng, Ailing and Li, Zhe and Zhang, Tianke and Yuan, Chun and Li, Yu},
  booktitle={ICCV},
  pages={17185--17194},
  year={2023}
}

@article{li2022asymmetric,
  title={Asymmetric temperature scaling makes larger networks teach well again},
  author={Li, Xin-Chun and Fan, Wen-Shu and Song, Shaoming and Li, Yinchuan and Yunfeng, Shao and Zhan, De-Chuan and others},
  journal={NeurIPS},
  volume={35},
  pages={3830--3842},
  year={2022}
}

@article{huang2022knowledge,
  title={Knowledge distillation from a stronger teacher},
  author={Huang, Tao and You, Shan and Wang, Fei and Qian, Chen and Xu, Chang},
  journal={NeurIPS},
  volume={35},
  pages={33716--33727},
  year={2022}
}

@inproceedings{guo2023class,
  title={Class Attention Transfer Based Knowledge Distillation},
  author={Guo, Ziyao and Yan, Haonan and Li, Hui and Lin, Xiaodong},
   booktitle={CVPR},
  pages={11868--11877},
  year={2023}
}

@inproceedings{romero2014fitnets,
  title={Fitnets: Hints for thin deep nets},
  author={Romero, Adriana and Ballas, Nicolas and Kahou, Samira Ebrahimi and Chassang, Antoine and Gatta, Carlo and Bengio, Yoshua},
   booktitle={ICLR},
  year={2015}
}

@inproceedings{chen2021distilling,
  title={Distilling knowledge via knowledge review},
  author={Chen, Pengguang and Liu, Shu and Zhao, Hengshuang and Jia, Jiaya},
  booktitle={CVPR},
  pages={5008--5017},
  year={2021}
}

@inproceedings{park2019relational,
  title={Relational knowledge distillation},
  author={Park, Wonpyo and Kim, Dongju and Lu, Yan and Cho, Minsu},
  booktitle={CVPR},
  pages={3967--3976},
  year={2019}
}

@article{niu2022respecting,
  title={Respecting transfer gap in knowledge distillation},
  author={Niu, Yulei and Chen, Long and Zhou, Chang and Zhang, Hanwang},
  journal={NeurIPS},
  volume={35},
  pages={21933--21947},
  year={2022}
}

@inproceedings{jin2023multi,
  title={Multi-level logit distillation},
  author={Jin, Ying and Wang, Jiaqi and Lin, Dahua},
  booktitle={CVPR},
  pages={24276--24285},
  year={2023}
}

@inproceedings{luo2024scale,
  title={Scaled Decoupled Distillation},
  author={Wei, Shicai and Luo, Chunbo and Luo, Yang},
  booktitle={CVPR},
  pages={15975--15983},
  year={2024}
}

@inproceedings{Sun2024Logit,
    title={Logit Standardization in Knowledge Distillation},
    author={Sun, Shangquan and Ren, Wenqi and Li, Jingzhi and Wang, Rui and Cao, Xiaochun},
    booktitle={CVPR},
  pages={15731--15740},
    year={2024}
}

@inproceedings{komodakis2017paying,
  title={Paying more attention to attention: improving the performance of convolutional neural networks via attention transfer},
  author={Komodakis, Nikos and Zagoruyko, Sergey},
  booktitle={ICLR},
  year={2017}
}

@inproceedings{tian2019contrastive,
  title={Contrastive representation distillation},
  author={Tian, Yonglong and Krishnan, Dilip and Isola, Phillip},
   booktitle={ICLR},
  year={2020}
}

@inproceedings{
liu2023functionconsistent,
title={Function-Consistent Feature Distillation},
author={Dongyang Liu and Meina Kan and Shiguang Shan and Xilin CHEN},
booktitle={ICLR},
year={2023},
}

@inproceedings{li2023curriculum,
  title={Curriculum temperature for knowledge distillation},
  author={Li, Zheng and Li, Xiang and Yang, Lingfeng and Zhao, Borui and Song, Renjie and Luo, Lei and Li, Jun and Yang, Jian},
  booktitle={AAAI},
  pages={1504--1512},
  year={2023}
}

@inproceedings{
zheng2024knowledge,
title={Knowledge Distillation Based on Transformed Teacher Matching},
author={Kaixiang Zheng and EN-HUI Yang},
booktitle={ICLR},
year={2024},
url={https://openreview.net/forum?id=MJ3K7uDGGl}
}

@article{xu2024improving,
  author={Xu, Liuchi and Ren, Jin and Huang, Zhenhua and Zheng, Weishi and Chen, Yunwen},
  journal={TCSVT}, 
  title={Improving Knowledge Distillation via Head and Tail Categories}, 
  year={2024},
  volume={34},
  number={5},
  pages={3465-3480},
 }

@article{krizhevsky2009learning,
  title={Learning multiple layers of features from tiny images},
  author={Krizhevsky, Alex and Hinton, Geoffrey and others},
  year={2009},
  journal={Toronto, ON, Canada}
}

@inproceedings{deng2009imagenet,
  title={Imagenet: A large-scale hierarchical image database},
  author={Deng, Jia and Dong, Wei and Socher, Richard and Li, Li-Jia and Li, Kai and Fei-Fei, Li},
  booktitle={CVPR},
  pages={248--255},
  year={2009}
}

@inproceedings{simonyan2014very,
  title={Very deep convolutional networks for large-scale image recognition},
  author={Simonyan, Karen and Zisserman, Andrew},
  booktitle = {ICLR},
  year={2015}
}

@inproceedings{howard2017mobilenets,
  title={Mobilenets: Efficient convolutional neural networks for mobile vision applications},
  author={Howard, Andrew G and Zhu, Menglong and Chen, Bo and Kalenichenko, Dmitry and Wang, Weijun and Weyand, Tobias and Andreetto, Marco and Adam, Hartwig},
  booktitle={CVPR},
  pages={484--492},
  year={2017}
}

@article{zou2024segment,
  title={Segment everything everywhere all at once},
  author={Zou, Xueyan and Yang, Jianwei and Zhang, Hao and Li, Feng and Li, Linjie and Wang, Jianfeng and Wang, Lijuan and Gao, Jianfeng and Lee, Yong Jae},
  journal={NeurIPS},
  volume={36},
pages = {19769--19782},
  year={2024}
}

@inproceedings{peng2019correlation,
  title={Correlation congruence for knowledge distillation},
  author={Peng, Baoyun and Jin, Xiao and Liu, Jiaheng and Li, Dongsheng and Wu, Yichao and Liu, Yu and Zhou, Shunfeng and Zhang, Zhaoning},
  booktitle={ICCV},
  pages={5007--5016},
  year={2019}
}

@inproceedings{heo2019comprehensive,
  title={A comprehensive overhaul of feature distillation},
  author={Heo, Byeongho and Kim, Jeesoo and Yun, Sangdoo and Park, Hyojin and Kwak, Nojun and Choi, Jin Young},
  booktitle={ICCV},
  pages={1921--1930},
  year={2019}
}

@inproceedings{yang2021knowledge,
  title={Knowledge distillation via softmax regression representation learning},
  author={Yang, Jing and Martinez, Brais and Bulat, Adrian and Tzimiropoulos, Georgios and others},
  year={2021},
  booktitle={ICLR}
}

@inproceedings{yang2022masked,
  title={Masked generative distillation},
  author={Yang, Zhendong and Li, Zhe and Shao, Mingqi and Shi, Dachuan and Yuan, Zehuan and Yuan, Chun},
  booktitle={ECCV},
  pages={53--69},
  year={2022}
}

@inproceedings{yang2022focal,
  title={Focal and global knowledge distillation for detectors},
  author={Yang, Zhendong and Li, Zhe and Jiang, Xiaohu and Gong, Yuan and Yuan, Zehuan and Zhao, Danpei and Yuan, Chun},
  booktitle={CVPR},
  pages={4643--4652},
  year={2022}
}

@article{wah2011caltech,
  title={The caltech-ucsd birds-200-2011 dataset},
  author={Wah, Catherine and Branson, Steve and Welinder, Peter and Perona, Pietro and Belongie, Serge},
journal={CNS-TR-2011-001},
  year={2011},
  publisher={California Institute of Technology}
}

@inproceedings{huang2024knowledge,
 author = {Huang, Tao and Zhang, Yuan and Zheng, Mingkai and You, Shan and Wang, Fei and Qian, Chen and Xu, Chang},
 booktitle = {NeurIPS},
 pages = {65299--65316},
 title = {Knowledge Diffusion for Distillation},
 volume = {36},
 year = {2023}
}

@inproceedings{
kim2024do,
title={Do Topological Characteristics Help in Knowledge Distillation?},
author={Jungeun Kim and Junwon You and Dongjin Lee and Ha Young Kim and Jae-Hun Jung},
booktitle={ICML},
year={2024},
}

@inproceedings{
liu2023norm,
title={{NORM}: Knowledge Distillation via N-to-One Representation Matching},
author={Xiaolong Liu and LUKING LI and Chao Li and Anbang Yao},
booktitle={ICLR},
year={2023},
}

@inproceedings{wang2019distilling,
  title={Distilling object detectors with fine-grained feature imitation},
  author={Wang, Tao and Yuan, Li and Zhang, Xiaopeng and Feng, Jiashi},
  booktitle={CVPR},
  pages={4933--4942},
  year={2019}
}

@InProceedings{Auxiliary,
  title = 	 {Knowledge Distillation with Auxiliary Variable},
  author = {Peng, Bo and Fang, Zhen and Zhang, Guangquan and Lu, Jie},
  booktitle ={ICML},
  pages = {40185--40199},
  year = 	 {2024},
  volume = 	 {235},
  series = 	 {Proceedings of Machine Learning Research},
  month = 	 {21--27 Jul},
  publisher =    {PMLR},
}

@inproceedings{ma2018shufflenet,
  title={Shufflenet v2: Practical guidelines for efficient cnn architecture design},
  author={Ma, Ningning and Zhang, Xiangyu and Zheng, Hai-Tao and Sun, Jian},
  booktitle={ECCV},
  pages={116--131},
  year={2018}
}

@article{hao2023one,
  title={One-for-all: Bridge the gap between heterogeneous architectures in knowledge distillation},
  author={Hao, Zhiwei and Guo, Jianyuan and Han, Kai and Tang, Yehui and Hu, Han and Wang, Yunhe and Xu, Chang},
  journal={NeurIPS},
  volume={36},
  pages={79570--79582},
  year={2023}
}

@inproceedings{
hossain2025single,
title={Single Teacher, Multiple Perspectives: Teacher Knowledge Augmentation for Enhanced Knowledge Distillation},
author={Md Imtiaz Hossain and Sharmen Akhter and Choong Seon Hong and Eui-Nam Huh},
booktitle={ICLR},
year={2025},
}

@inproceedings{li2024detkds,
  title={Detkds: Knowledge distillation search for object detectors},
  author={Li, Lujun and Bao, Yufan and Dong, Peijie and Yang, Chuanguang and Li, Anggeng and Luo, Wenhan and Liu, Qifeng and Xue, Wei and Guo, Yike},
  booktitle={ICML},
  year={2024}
}

@article{lin2025feature,
    author    = {Lin, Jhe-Hao and Yao, Yi and Hsu, Chan-Feng and Xie, Hong-Xia and Shuai, Hong-Han and Cheng, Wen-Huang},
    title   = {Perspective-Aware Teaching: Adapting Knowledge for Heterogeneous Distillation},
    journal = {ICCV},
    year      = {2025},
    pages     = {4178-4187}
}

@inproceedings{zheng2025hierarchical,
  title={Hierarchical cross-modal prompt learning for vision-language models},
  author={Zheng, Hao and Yang, Shunzhi and He, Zhuoxin and Yang, Jinfeng and Huang, Zhenhua},
  booktitle={ICCV},
  pages={1891--1901},
  year={2025}
}

@article{dosovitskiy2020image,
  title={An image is worth 16x16 words: Transformers for image recognition at scale},
  author={Dosovitskiy, Alexey and Beyer, Lucas and Kolesnikov, Alexander and Weissenborn, Dirk and Zhai, Xiaohua and Unterthiner, Thomas and Dehghani, Mostafa and Minderer, Matthias and Heigold, Georg and Gelly, Sylvain and others},
  journal={ICLR},
  year={2021}
}

@article{tolstikhin2021mlp,
  title={Mlp-mixer: An all-mlp architecture for vision},
  author={Tolstikhin, Ilya O and Houlsby, Neil and Kolesnikov, Alexander and Beyer, Lucas and Zhai, Xiaohua and Unterthiner, Thomas and Yung, Jessica and Steiner, Andreas and Keysers, Daniel and Uszkoreit, Jakob and others},
  journal={NeurIPS},
  volume={34},
  pages={24261--24272},
  year={2021}
}

@inproceedings{wang2025debiased,
  title={Debiased Distillation for Consistency Regularization},
  author={Wang, Lu and Xu, Liuchi and Yang, Xiong and Huang, Zhenhua and Cheng, Jun},
  booktitle={AAAI},
  pages={7799--7807},
  year={2025}
}

@inproceedings{zhou2025all,
  title={All You Need in Knowledge Distillation Is a Tailored Coordinate System},
  author={Zhou, Junjie and Zhu, Ke and Wu, Jianxin},
  booktitle={AAAI},
  pages={22946--22954},
  year={2025}
}

@article{li2024tas,
  title={TAS: Distilling Arbitrary Teacher and Student via a Hybrid Assistant},
  author={Li, Guopeng and Wang, Qiang and Yan, Ke and Ding, Shouhong and Gao, Yuan and Xia, Gui-Song},
  journal={arXiv preprint arXiv:2410.12342},
  year={2024}
}

@inproceedings{huang2025distilling,
  title={Distilling Knowledge from Heterogeneous Architectures for Semantic Segmentation},
  author={Huang, Yanglin and Hu, Kai and Zhang, Yuan and Chen, Zhineng and Gao, Xieping},
  booktitle={AAAI},
  pages={3824--3832},
  year={2025}
}

@inproceedings{liu2021swin,
  title={Swin transformer: Hierarchical vision transformer using shifted windows},
  author={Liu, Ze and Lin, Yutong and Cao, Yue and Hu, Han and Wei, Yixuan and Zhang, Zheng and Lin, Stephen and Guo, Baining},
  booktitle={ICCV},
  pages={10012--10022},
  year={2021}
}

@inproceedings{touvron2021training,
  title={Training data-efficient image transformers \& distillation through attention},
  author={Touvron, Hugo and Cord, Matthieu and Douze, Matthijs and Massa, Francisco and Sablayrolles, Alexandre and J{\'e}gou, Herv{\'e}},
  booktitle={ICML},
  pages={10347--10357},
  year={2021},
  organization={PMLR}
}

@inproceedings{zhao2023cumulative,
  title={Cumulative spatial knowledge distillation for vision transformers},
  author={Zhao, Borui and Song, Renjie and Liang, Jiajun},
  booktitle={ICCV},
  pages={6146--6155},
  year={2023}
}

@article{lee2025customkd,
  title={CustomKD: Customizing Large Vision Foundation for Edge Model Improvement via Knowledge Distillation},
  author={Lee, Jungsoo and Das, Debasmit and Hayat, Munawar and Choi, Sungha and Hwang, Kyuwoong and Porikli, Fatih},
  journal={CVPR},
  year={2025}
}

@inproceedings{liu2022cross,
  title={Cross-architecture knowledge distillation},
  author={Liu, Yufan and Cao, Jiajiong and Li, Bing and Hu, Weiming and Ding, Jingting and Li, Liang},
  booktitle={ACCV},
  pages={3396--3411},
  year={2022}
}

@article{sun2024knowledge,
  title={Knowledge distillation with refined logits},
  author={Sun, Wujie and Chen, Defang and Lyu, Siwei and Chen, Genlang and Chen, Chun and Wang, Can},
  journal={ICCV },
  year={2025}
}

@article{
oquab2024dinov,
title={{DINO}v2: Learning Robust Visual Features without Supervision},
author={Maxime Oquab and Timoth{\'e}e Darcet and Th{\'e}o Moutakanni and Huy V. Vo and Marc Szafraniec and Vasil Khalidov and Pierre Fernandez and Daniel HAZIZA and Francisco Massa and Alaaeldin El-Nouby and Mido Assran and Nicolas Ballas and Wojciech Galuba and Russell Howes and Po-Yao Huang and Shang-Wen Li and Ishan Misra and Michael Rabbat and Vasu Sharma and Gabriel Synnaeve and Hu Xu and Herve Jegou and Julien Mairal and Patrick Labatut and Armand Joulin and Piotr Bojanowski},
journal={Transactions on Machine Learning Research},
issn={2835-8856},
year={2024},
}

@article{tschannen2025siglip,
  title={Siglip 2: Multilingual vision-language encoders with improved semantic understanding, localization, and dense features},
  author={Tschannen, Michael and Gritsenko, Alexey and Wang, Xiao and Naeem, Muhammad Ferjad and Alabdulmohsin, Ibrahim and Parthasarathy, Nikhil and Evans, Talfan and Beyer, Lucas and Xia, Ye and Mustafa, Basil and others},
  journal={arXiv preprint arXiv:2502.14786},
  year={2025}
}

@article{xu2025local,
    title={Local Dense Logit Relations for Enhanced Knowledge Distillation},
    author={Liuchi Xu and Kang Liu and Jinshuai Liu and Lu Wang and Lisheng Xu and Jun Cheng},
    year={2025},
    journal={ICCV},
}

@article{guan2025enhancing,
    title={Enhancing Logits Distillation with Plug Play Kendall's $\tau$ Ranking Loss},
    author={Yuchen Guan and Runxi Cheng and Kang Liu and Chun Yuan},
    year={2025},
    journal={ICML},
}

@article{wu2024aligning,
      title={Aligning in a Compact Space: Contrastive Knowledge Distillation between Heterogeneous Architectures}, 
      author={Hongjun Wu and Li Xiao and Xingkuo Zhang and Yining Miao},
      year={2024},
      journal={arXiv preprint arXiv:2405.2405},
}

@article{yang2025multilevel,
      title={Multi-Level Decoupled Relational Distillation for Heterogeneous Architectures}, 
      author={Yaoxin Yang and Peng Ye and Weihao Lin and Kangcong Li and Yan Wen and Jia Hao and Tao Chen},
      year={2025},
      journal={arXiv preprint arXiv:2502.06189},
}

@article{maji2013fine,
  title={Fine-grained visual classification of aircraft},
  author={Maji, Subhransu and Rahtu, Esa and Kannala, Juho and Blaschko, Matthew and Vedaldi, Andrea},
  journal={arXiv preprint arXiv:1306.5151},
  year={2013}
}

@article{zhang2025cross,
      title={Cross-Architecture Distillation Made Simple with Redundancy Suppression}, 
      author={Weijia Zhang and Yuehao Liu and Wu Ran and Chao Ma},
      year={2025},
    journal={ICCV}, 
}

@inproceedings{kornblith2019similarity,
  title={Similarity of neural network representations revisited},
  author={Kornblith, Simon and Norouzi, Mohammad and Lee, Honglak and Hinton, Geoffrey},
  booktitle={ICML},
  pages={3519--3529},
  year={2019},
  organization={PMlR}
}

@article{cortes2012algorithms,
  title={Algorithms for learning kernels based on centered alignment},
  author={Cortes, Corinna and Mohri, Mehryar and Rostamizadeh, Afshin},
  journal={The Journal of Machine Learning Research},
  volume={13},
  number={1},
  pages={795--828},
  year={2012},
}

@article{li2022shadow,
  title={Shadow knowledge distillation: Bridging offline and online knowledge transfer},
  author={Li, Lujun and Jin, Zhe},
  journal={NeuIPS},
  volume={35},
  pages={635--649},
  year={2022}
}

@article{li2023kd,
  title={Kd-zero: Evolving knowledge distiller for any teacher-student pairs},
  author={Li, Lujun and Dong, Peijie and Li, Anggeng and Wei, Zimian and Yang, Ya},
  journal={NeuIPS},
  volume={36},
  pages={69490--69504},
  year={2023}
}

@inproceedings{li2023automated,
  title={Automated knowledge distillation via monte carlo tree search},
  author={Li, Lujun and Dong, Peijie and Wei, Zimian and Yang, Ya},
  booktitle={ICCV},
  pages={17413--17424},
  year={2023}
}

@inproceedings{dong2023diswot,
  title={Diswot: Student architecture search for distillation without training},
  author={Dong, Peijie and Li, Lujun and Wei, Zimian},
  booktitle={CVPR},
  pages={11898--11908},
  year={2023}
}

\clearpage
\newpage
\appendix

% New page for the appendix title
\twocolumn[{%
    \begin{center}
        \LARGE  \textbf{Supplementary Materials for Heterogeneous Complementary Distillation} 
    \end{center}
    \vspace{1cm}
}]
\subsection{Optimization Objective}
The algorithm for implementing the HCD method is shown in pseudo code in Algorithm~\ref{alg}.

\begin{algorithm}[!ht]
\caption{Pseudo code for HCD}
\begin{algorithmic}[1]
\Require 
    $\mathcal{D}_{set}$: training dataset, $\mathbf{x}$ and $\mathbf{y}$ represent sample and one-hot label, respectively; 
$f_{\text{net}}^{t}$: teacher network; 
$f_{\text{net}}^{s}$: student network with parameters $\theta$; 
    $\epsilon$: learning rate; $\lambda$, $\beta$, and $\omega$ are the weight coefficients.
\Ensure 
Trained parameters $\theta$ of the student network $f_{\text{net}}^{s}$
\State Randomly initialize parameters $\theta$ of the student network $f_{\text{net}}^{s}$ and  loading teacher network $f_{\text{net}}^{t}$.
\Repeat
    \State Iterative updates a mini-batch $\mathcal{B}$ from $\mathcal{D}_{set}$
    \For{each $(\mathbf{x}, \mathbf{y}) \in \mathcal{B}$}
        \State $\mathbf{z}^{t} \leftarrow f_{\text{net}}^{t}(\mathbf{x})$ \Comment{Compute teacher's output via teacher network $f_{\text{net}}^{t}$ for sample $\mathbf{x}$}.
        \State $\mathbf{z}^{s} \leftarrow f_{\text{net}}^{s}(\mathbf{x})$ \Comment{Compute student's output via student network $f_{\text{net}}^{s}$ for sample $\mathbf{x}$}.
        \State (1) Compute the task loss $\mathcal{L}_{\text{CE}}$ (i.e., cross entropy) via sample $\mathbf{x}$ and $\mathbf{y}$, and $\mathcal{L}_{\text{KL}}$ via $\mathbf{z}^{s}$ and $\mathbf{z}^{t}$. 
        \State (2) Extract shared logits using a projector and decouple it to obtain multiple sub-logits using Eq.~(\ref{cfm1}).
        \State (3) Transfer knowledge from sub-logit via $\mathcal{L}_{KD}^{sub}$ in Eq.(~\ref{kd}) and $\mathcal{L}_{CE}^{sub}$ in Eq.(~\ref{ce_loss})
        \State (4) Orthogonal constraints ensure the diversity of sub-logit $\mathcal{L}_{orth}$ in Eq.~(\ref{orth})
    \EndFor
    \State $\mathcal{L}_{HCD} \leftarrow \dfrac{1}{|\mathcal{B}|}( \mathcal{L}_{CE
}+\mathcal{L}_{CE}^{sub}+ \lambda\mathcal{L}_{KL}+\beta \mathcal{L}_{KL}^{sub}+\omega \mathcal{L}_{orth
})$
    \State Update $\theta \leftarrow \theta - \epsilon \left( \nabla_{\theta} \mathcal{L}_{HCD} \right)$
\Until{The maximum number of iterations is reached}
\State \Return $\theta$
\end{algorithmic}
\label{alg}
\end{algorithm}
\vspace{-10pt}
\subsection{Experiment settings}
\textbf{Baselines.} To show the performance of our proposed HCD framework, we benchmark it against leading knowledge distillation techniques. In particular, for feature-based distillation, we choose FitNets~\cite{romero2014fitnets}, AT~\cite{komodakis2017paying}, ReviewKD~\cite{chen2021distilling}, OFD~\cite{heo2019comprehensive}, FCFD~\cite{liu2023functionconsistent}, CAT-KD~\cite{guo2023class}, RKD~\cite{park2019relational}, CC~\cite{peng2019correlation}, CRD~\cite{tian2019contrastive}, RSD~\cite{zhang2025cross}, and PAT~\cite{lin2025feature} . For logit-based distillation, we choose KD~\cite{hinton2015distilling}, DKD~\cite{zhao2022decoupled}, IPWD~\cite{niu2022respecting},
CTKD~\cite{li2023curriculum},
SDD~\cite{luo2024scale},  LSKD~\cite{Sun2024Logit}, DIST~\cite{huang2022knowledge}, OFA-KD~\cite{hao2023one}, WTTM~\cite{zheng2024knowledge}, RKKD~\cite{guan2025enhancing}, LDRLD~\cite{xu2025local}, RLD~\cite{sun2024knowledge}, and TeKAP~\cite{hossain2025single}.

\textbf{Models.} To evaluate the effectiveness of our HCD framework in heterogeneous knowledge distillation, we employ various teacher-student architectures, including Convolutional Neural Networks (CNNs), Vision Transformers (ViTs) and MLP-Mixers. Specifically, we evaluate the following architectures on the CIFAR-100 and ImageNet-1K: ResNet-50~\cite{he2016deep},ViT~\cite{dosovitskiy2020image}, Swin-Tiny~\cite{liu2021swin}, MLP-Mixer-B/16~\cite{tolstikhin2021mlp}, DeiT-Tiny~\cite{touvron2021training}, ResNet-18~\cite{he2016deep}, and  MobileNetV1~\cite{howard2017mobilenets}, with larger models (e.g., ResNet-50, Swin-Tiny) as teachers and smaller models (e.g., ResNet-18, MobileNetV1) as students.
For training fine-grained datasets, we use the encoder of the visual base model as the teacher model, specifically ViT-L and ViT-S from DINOv2~\cite{oquab2024dinov}, ViT-L and ViT-B from SigLIP2~\cite{tschannen2025siglip}. We use some CNNs as student model, such as MobileNetV2~\cite{sandler2018mobilenetv2}, VGG~\cite{simonyan2014very}, ResNet20, ShuffleNetV2~\cite{ma2018shufflenet}.

\textbf{Training details:} (1) We employ different optimizers for various student model architectures. Specifically, we train all CNN student models (e.g., ResNet-18, MobileNetV2) using SGD, while for Vision Transformers (ViTs) (e.g., Swin-Tiny) and MLP-Mixers (e.g., MLP-Mixer-B/16), we use the AdamW optimizer. We train all models for 300 epochs on the CIFAR-100 dataset and we train CNN students for 100 epochs on the  ImageNet-1K dataset. For more detailed parameter configuration, see OFA-KD~\cite{hao2023one}.  (2) The batch size of 256 is used for same-architecture on the ImageNet-1K dataset, and the initial learning rate is set to 0.1. A linear warm-up is applied during the first 10 epochs. The learning rate is then reduced by a factor of 10 at the 30th, 60th, and 90th epochs. For more detailed parameter configuration, see DKD~\cite{zhao2022decoupled}. 
For the student model ResNet18 on the CIFAR100 dataset, the settings are $\lambda$=1.0, $\omega$=10.0, $\beta$=8.0, and temperature $\tau$=4.0. For the student model MobileNetV2 on the CIFAR100 dataset, the settings are $\lambda$=1.0, $\omega$=1.0,  $\beta$=2.0, and temperature $\tau$=4.0. For the parameter settings of heterogeneous distillation on ImageNet-1K, see Table~\ref{imagenet_settings} and same architecture distillation on ImageNet-1K, see Table~\ref{Hyper_imagenet}. (3) We fine-tune the linear classification heads of DINOv2 and SigLIP2 on the CUB-200-2011 and FGVC-Aircraft datasets for 20 epochs, with a batch size of 128 and a learning rate of 0.1, serving as the teacher models (ViT-S,ViT-L,ViT-B). The detailed parameter settings for training the student models are similar to those used in CRD~\cite{tian2019contrastive} on the CIFAR100 dataset.
\begin{table}[!ht]
\centering
\setlength\tabcolsep{0.75pt}
\scalebox{0.78}{
\begin{tabular}{c|ccccc}
\tophline
\begin{tabular}{c} 
Teacher \\ Student
\end{tabular}  & 
\begin{tabular}{c} 
Swin-T \\ ResNet18
\end{tabular} & 
\begin{tabular}{c} 
Mixer-B/16 \\ ResNet18
\end{tabular} & 
\begin{tabular}{c} 
DeiT-T \\ MobileNetV2
\end{tabular} & 
\begin{tabular}{c} 
Swin-T \\ MobileNetV2
\end{tabular} & 
\begin{tabular}{c} 
Mixer-B/16 \\ MobileNetV2
\end{tabular}\\
\hline
\hline
\multirow{3}*{HCD}&$\lambda$=1.0  & $\lambda$=1.0 & $\lambda$=1.0 &$\lambda$=1.0&$\lambda$=1.0 \\
&$\beta$=8.0  &$\beta$=8.0  &$\beta$=2.0&$\beta$=2.0&$\beta$=2.0\\
&$\omega$=10.0 & $\omega$=10.0 &$\omega$=2.0&$\omega$=2.0&$\omega$=2.0 \\
\tophline
\end{tabular}
}
\vspace{-6pt}
\caption{Hyperparameters for Heterogeneous architecture distillation on ImageNet-1K dataset.}
\label{imagenet_settings}
\end{table}
\begin{table}[!ht]
\vspace{-6pt}
\centering 
\setlength\tabcolsep{10.0pt}
\scalebox{0.9}{% Adjust the width to fit the content better
\begin{tabular}{c|c|cc}
\tophline
\multirow{2}*{Method}&Teacher & ResNet34& ResNet50 \\ 
& Student & ResNet18 & MobileNetV1 \\
\hline
\hline
\multirow{3}*{HCD}&$\lambda$  & 8.0 & 8.0 \\
&$\beta$  & 2.0 & 2.0 \\
&$\omega$  & 2.0 & 2.0 \\
\tophline
\end{tabular}
}
\vspace{-6pt}
\caption{Hyperparameters for homogeneous  architecture distillation on the ImageNet-1K dataset.}
\label{Hyper_imagenet}
\vspace{-12pt}
\end{table}
\subsection{Ablation Studies}
\textbf{Impact of training cost.} We conducted the ablation study on CIFAR-100 with Swin-T as the teacher and ResNet18 as the student, which demonstrates that our proposed HCD achieves a Top-1 accuracy of 82.78\% in Table~\ref{table:training efficiency}, surpassing KD (78.74\%), OFA (80.54\%), and LDRLD (82.17\%). However, HCD incurs higher computational costs, with 14.32 seconds per epoch and 14.39 GB memory cost per GPU, compared to KD’s 12.26 seconds and 13.18 GB. This trade-off is justified by HCD’s superior performance, leveraging the CFM across all feature stages to enhance knowledge transfer in heterogeneous knowledge distillation.
\begin{table}[!ht]
\centering
\resizebox{0.99\linewidth}{!}{
\begin{tabular}{ccccc}
\tophline
 Evaluation Metrics& KD & OFA & LDRLD &  HCD  \\
 \midhline
Time (s/epoch) & \bf 12.26 & 13.38 & 12.58 & 14.32  \\ 
Memory (GB/GPU) & \bf 13.18 & 14.19 & 13.38 & 14.39  \\ 
Top-1 Acc (\%) & 78.74 & 80.54 & 82.17 & \bf 82.78 \\ 
\bottomline
\end{tabular}
}
\caption{Ablation study of training cost on the CIFAR-100 dateset with 8 NVIDIA RTX 3090 GPUs for Swin-T (teacher) and ResNet18 (student).}
\label{table:training efficiency}
\vspace{-0.3cm}
\end{table}

\noindent\textbf{Impact of orthogonal loss for threshold $\theta$.} We analyze the impact of the threshold $\theta$ in our orthogonality loss, with results presented in Table~\ref{table:theta}. The results indicate that our method is robust to the choice of $\theta$, as performance remains stable across different values for all tested teacher-student pairs. Notably, $\theta$=0.5 not only achieves the highest accuracy (82.78\%) in the Swin-T vs. ResNet18 setting but also maintains highly competitive performance in all other configurations (e.g., Swin-T vs. MobileNetV2 and ViT-S vs. MobileNetV2). This demonstrates that it provides a well-balanced constraint between encouraging sub-logit diversity and preserving stable knowledge transfer. Therefore, we adopt $\theta$= 0.5 as the default value for all our experiments.
\begin{table}[!ht]
\centering
\resizebox{0.99\linewidth}{!}{
\begin{tabular}{ccccc}
\tophline
\begin{tabular}{c} 
\multirow{1}{*}{threshold} 
\end{tabular}  & 
\begin{tabular}{c} 
Swin-T \\ ResNet18
\end{tabular} & 
\begin{tabular}{c} 
ViT-S \\ ResNet18
\end{tabular} & 
\begin{tabular}{c} 
Swin-T \\ MobileNetV2
\end{tabular} & 
\begin{tabular}{c} 
ViT-S \\ MobileNetV2
\end{tabular} 
\\
 \midhline
$\theta$=0.25 &  82.46 & 81.61 & 82.20 & 80.80  \\ 
$\theta$=0.50 &   82.78  & 81.33 & 82.19 & 80.81  \\ 
$\theta$=0.75 &  82.63 & 81.74 & 82.12 &  81.02 \\ 
\bottomline
\end{tabular}
}
\caption{Ablation study of $\theta$ on the CIFAR-100 dateset.}
\label{table:theta}
\vspace{-0.3cm}
\end{table}

\textbf{Impact of the number of student's stages.} We fused the low-level features of the student with the high-level features of the teacher to facilitate the transfer of complementary knowledge. As shown in Table~\ref{Impact of the stages}, this approach leads to significant performance improvements with an increasing number of student stages, suggesting that  high-level abstract features enhance the discrimination of low-level features. These results demonstrate the effectiveness of the teacher-student knowledge complementarity.
\vspace{-4pt}
\begin{table}[!ht]
\centering
\begin{tabular}{c|c}
\tophline
Stages Used & Top-1 Accuracy (\%) \\
\midhline
None & 74.01 \\
\{1\} & 81.82 (+7.81) \\
\{1, 2\} & 82.34 (+8.33) \\
\{1, 2, 3\} & 82.46 (+8.45) \\
\{1, 2, 3, 4\} & \textbf{82.78} (+8.77) \\
\bottomline
\end{tabular}
\caption{Ablation study of the number of stages on student’s performance with HCD on the CIFAR100 dataset (Swin-Tiny as a teacher, ResNet18 as a student).}
\label{Impact of the student's stages}
\end{table}

\textbf{Impact of the number of student's stages.} We fused the low-level features of the student with the high-level features of the teacher to facilitate the transfer of complementary knowledge. As shown in Table~\ref{Impact of the student's stages}, this approach leads to significant performance improvements with an increasing number of student stages, suggesting that  high-level abstract features enhance the discrimination of low-level features. These results demonstrate the effectiveness of the teacher-student knowledge complementarity.
\vspace{-4pt}
\begin{table}[!ht]
\centering
\label{tab:stage_ablation}
\begin{tabular}{c|c}
\tophline
Stages Used & Top-1 Accuracy (\%) \\
\midhline
None & 74.01 \\
\{1\} & 81.82 (+7.81) \\
\{1, 2\} & 82.34 (+8.33) \\
\{1, 2, 3\} & 82.46 (+8.45) \\
\{1, 2, 3, 4\} & \textbf{82.78} (+8.77) \\
\bottomline
\end{tabular}
\caption{Ablation study of the number of stages on student’s performance with HCD on the CIFAR100 dataset (Swin-Tiny as a teacher, ResNet18 as a student).}
\label{Impact of the stages}
\end{table}

\subsection{Discussion of Fusion Knowledge.}
(1) \textbf{Fusion Ratio.} We rewrite the Eq.~(\ref{fusion}) to get the following expression:
\begin{equation}
\begin{aligned}
\mathbf{z}_{i}^{j} \gets \lambda_1* \mathbf{z}_{i}^{j} \oplus  \lambda_2* \mathbf{z}^t,
\end{aligned}
\label{fusion1}
\end{equation}
where $\lambda_1$ + $\lambda_2$ =1. The fusion ratio between teacher and student knowledge presents a critical trade-off. As shown in Table~\ref{ratio}, while reducing the teacher's contribution helps mitigate overfitting, insufficient guidance can cause the learning process to drift from task objectives. Conversely, excessive reliance on the teacher, similar to methods like OFA, suppresses knowledge diversity and leads to suboptimal performance.
\vspace{-4pt}
\begin{table}[!ht]
\centering
\begin{tabular}{cc|c}
\tophline
$\lambda_1$&$\lambda_2$ & Top-1 Accuracy (\%) \\
\midhline
None &None& 74.01 \\
0.25&0.75& 81.84 (+7.83) \\
0.50 &0.50& 81.96 (+7.95) \\
0.75 &0.25& 82.06 (+8.05) \\
\bottomline
\end{tabular}
\vspace{-4pt}
\caption{Ablation study of the fusion ratio on student’s performance with HCD on the CIFAR100 dataset (Swin-Tiny as a teacher, ResNet18 as a student).}
\label{ratio}
\vspace{-8pt}
\end{table}

\begin{table}[!ht]
\centering
\begin{tabular}{cc|c}
\tophline
$\lambda_3$&$\lambda_4$ & Top-1 Accuracy (\%) \\
\midhline
None &None& 74.01 \\
0.50 &0.50& 81.96 (+7.95) \\
0.75&0.75& 82.51 (+8.50) \\
1.0 &1.0& 82.78 (+8.77) \\
1.5 &1.5& 82.47 (+8.46) \\
\bottomline
\end{tabular}
\vspace{-4pt}
\caption{Ablation study of the weighted sum on student’s performance with HCD on the CIFAR100 dataset (Swin-Tiny as a teacher, ResNet18 as a student).}
\label{ratio2}
\end{table}

(2) \textbf{Weighted Sum.} We also rewrite the Eq.~(\ref{fusion}) to get the following expression:
\begin{equation}
\begin{aligned}
\mathbf{z}_{i}^{j} \gets \lambda_3* \mathbf{z}_{i}^{j} \oplus  \lambda_4* \mathbf{z}^t,
\end{aligned}
\label{partation}
\end{equation}
where $\lambda_3$ = $\lambda_4$. $\lambda_3$ and $\lambda_4$ are independent weights that control the contributions of $\mathbf{z}_{i}^{j}$ and $\mathbf{z}^t$, respectively. The weighted sum approach outperforms the fusion ratio method, primarily due to its flexible weight allocation, enhanced knowledge contribution, reduced reliance on a single source, optimized knowledge balance, and moderate mitigation of overfitting risks. The optimal performance is achieved at $\lambda_3$ = $\lambda_4$ = 1.0 (82.78\%), suggesting that moderate weight amplification is critical in Table~\ref{ratio2}. However, the decline performance observed at $\lambda_3$ = $\lambda_4$ = 1.5 = indicates the need for further tuning to prevent overfitting in Table~\ref{ratio2}.
\begin{table}[!ht]
\centering
\resizebox{0.99\linewidth}{!}{
\begin{tabular}{cc|cc}
\tophline
$\lambda_3$&$\lambda_4$ & Swin-T vs. ResNet18 & ViT-S vs. ResNet18 \\
\midhline
None &None& 74.01 &74.01\\
1.0 &0.0& 82.01 (+8.00) & 80.92 (+6.91) \\
1.0&1.0& 82.78 (+8.77)&81.33 (+7.32)\\
\bottomline
\end{tabular}
}
\vspace{-4pt}
\caption{Ablation study of the without teacher's logit on student’s performance with HCD on the CIFAR100 dataset (Swin-Tiny as a teacher, ResNet18 as a student).}
\label{ratio3}
\end{table}

(3) \textbf{Without fusing teacher's logit knowledge.} Furthermore, we conducted an ablation study on the fusion of teacher's logit knowledge, as shown in Table~\ref{ratio3}. The results demonstrate that incorporating the teacher's logits consistently improves performance, suggesting that this guidance effectively prevents the student from learning task-irrelevant knowledge and serves a crucial corrective role.
\end{document}